\def\year{2021}\relax
\documentclass[letterpaper]{article} % DO NOT CHANGE THIS
\usepackage{aaai21}  % DO NOT CHANGE THIS
\usepackage{times}   % DO NOT CHANGE THIS
\usepackage{helvet} % DO NOT CHANGE THIS
\usepackage{courier}  % DO NOT CHANGE THIS
\usepackage[hyphens]{url}  % DO NOT CHANGE THIS 
\usepackage{graphicx} % DO NOT CHANGE THIS
\usepackage{natbib}  % DO NOT CHANGE THIS AND DO NOT ADD ANY OPTIONS TO IT
\usepackage{caption} % DO NOT CHANGE THIS AND DO NOT ADD ANY OPTIONS TO IT
\nocopyright

\usepackage{amsfonts}
\usepackage{amsmath}
\usepackage{caption}
\usepackage{multirow}
\usepackage{diagbox}
\usepackage{bm}
\usepackage{subcaption}
\usepackage[algoruled,linesnumbered]{algorithm2e}
\usepackage{booktabs} 
\usepackage[switch]{lineno}

\usepackage{float} 
\newcommand{\method}{{DNIS}~}
\newcommand{\methodns}{{DNIS}}
\newcommand{\eat}[1]{}

\title{Differentiable Neural Input Search for Recommender Systems}

\author{Weiyu Cheng, Yanyan Shen, Linpeng Huang\\}
\affiliations {
Shanghai Jiao Tong University
\\
\{weiyu\_cheng, shenyy, lphuang\}@sjtu.edu.cn}
\begin{document}
%\linenumbers

\maketitle

%%\vspace{-0.15in}
\begin{abstract}
Latent factor models are the driving forces of the state-of-the-art recommender systems, with an important insight of vectorizing raw input features into dense embeddings. The dimensions of different feature embeddings are often set to a same value empirically, which limits the predictive performance of latent factor models. Existing works have proposed heuristic or reinforcement learning-based methods to search for mixed feature embedding dimensions. For efficiency concern, these methods typically choose embedding dimensions from a restricted set of candidate dimensions. However, this restriction will hurt the flexibility of dimension selection, leading to suboptimal performance of search results.

In this paper, we propose Differentiable Neural Input Search (DNIS), a method that searches for mixed feature embedding dimensions in a more flexible space through continuous relaxation and differentiable optimization. The key idea is to introduce a soft selection layer that controls the significance of each embedding dimension, and optimize this layer according to model's validation performance. DNIS is model-agnostic and thus can be seamlessly incorporated with existing latent factor models for recommendation. We conduct experiments with various architectures of latent factor models on three public real-world datasets for rating prediction, Click-Through-Rate (CTR) prediction, and top-k item recommendation. The results demonstrate that our method achieves the best predictive performance compared with existing neural input search approaches with fewer embedding parameters and less time cost.
\end{abstract}
 
\section{Introduction}

Most state-of-the-art recommender systems employ latent factor models that vectorize raw input features into dense embeddings. A key question often asked of feature embeddings is: ``How should we determine the dimensions of feature embeddings?'' The common practice is to set a uniform dimension for all the features, and treat the dimension as a hyperparameter that needs to be tuned with time-consuming grid search on a validation set. However, a uniform embedding dimension is not necessarily suitable for different features. 
Intuitively, predictive and popular features that appear in a large number of data samples usually deserve a high embedding dimension, which
encourages a high model capacity to fit these data samples~\cite{DBLP:conf/kdd/JoglekarLCXWAKL20,zhao2020autoemb}. 
Likewise, less predictive and infrequent features would rather be assigned with lower embedding dimensions to avoid overfitting on scarce data samples. 
As such, it is desirable to impose a mixed dimension scheme for different features towards better recommendation performance.
Another notable fact is that embedding layers in industrial web-scale recommender systems account for the majority of model parameters and can consume hundreds of gigabytes of memory space~\cite{park2018deep, DBLP:conf/recsys/CovingtonAS16}. 
Replacing a uniform feature embedding dimension with varying dimensions can help to remove redundant embedding weights and significantly reduce memory and storage costs for model parameters. 
Some recent works~\cite{ginart2019mixed,DBLP:conf/kdd/JoglekarLCXWAKL20} have focused on searching for mixed feature embedding dimensions automatically, which is defined as the \emph{Neural Input Search (NIS)} problem.
%The main idea of these works is to reduce the search space by choosing feature dimensions from a small set of empirically designed candidates. For example,
%Specifically, 
There are two existing approaches to dealing with NIS: \emph{heuristic-based} and \emph{reinforcement learning-based}. The heuristic-based approach~\cite{ginart2019mixed} designs an empirical function to determine the embedding dimensions for different features according to their frequencies of occurrence (also called popularity). The empirical function involves several hyperparameters that need to be manually tuned to yield a good result. 
The reinforcement learning-based approach~\cite{DBLP:conf/kdd/JoglekarLCXWAKL20}
% proposed a reinforcement learning-based method for addressing the NIS problem. They first 
 first divides a base dimension space equally into several blocks, and then applies reinforcement learning to generate decision sequences on the selection of dimension blocks for different features.
These methods, however, restrict each feature embedding dimension to be chosen from a small set of candidate dimensions that is explicitly predefined~\cite{DBLP:conf/kdd/JoglekarLCXWAKL20} or implicitly controlled by hyperparameters~\cite{ginart2019mixed}. 
Although this restriction reduces search space and thereby improves computational efficiency, another question then arises:  % it leads to another nontrivial question of 
how to decide the candidate dimensions? Notably, a suboptimal set of candidate dimensions could result in a suboptimal search result that hurts model's recommendation performance.

In this paper, we propose Differentiable Neural Input Search (\methodns) to address the NIS problem in a differentiable manner through gradient descent. Instead of searching over a predefined discrete set of candidate dimensions, \method relaxes the search space to be continuous and optimizes the selection of embedding dimensions by descending model's validation loss. More specifically, 
%We first sort features according to their frequency and divide them into blocks, where features in each block will share the same embedding dimension. 
we introduce a \emph{soft selection layer} between the embedding layer and the feature interaction layers of latent factor models. Each input feature embedding is fed into the soft selection layer to perform an element-wise multiplication with a scaling vector. 
The soft selection layer directly controls the significance of each dimension of the feature embedding, and it is essentially a part of model architecture which can be optimized according to model's validation performance. We also propose a gradient normalization technique to solve the problem of high variance of gradients when optimizing the soft selection layer.
%keep the backpropagated gradients steady during the optimization of the selection layer. 
%
After training, we employ a fine-grained pruning procedure by merging the soft selection layer with the feature embedding layer,
% to prune redundant or less informative embedding dimensions per feature, 
which yields a fine-grained mixed embedding dimension scheme for different features.
%, and the embeddings can then be pruned and formed into different dimensions. 
%\method is model-agnostic and 
\method can be seamlessly applied to various existing architectures of latent factor models for recommendation. We conduct extensive experiments with six existing architectures of latent factor model on three public real-world datasets for rating prediction, Click-Through-Rate (CTR) prediction, and top-k item recommendation tasks. The results demonstrate that the proposed method achieves the best predictive performance compared with existing NIS methods with fewer embedding parameters and less time cost.

% higher embedding dimension compression rates of over $2\times$ on CF task and over $20\times$ on CTR prediction task.
%We conduct experiments on two benchmark datasets of collaborative filtering and click-through rate prediction, where our method achieves the best performance and efficiency on different architectures with the least parameters.

The major contributions of this paper are summarized as follows:
\begin{itemize}
	\item We propose \methodns, a method that addresses the NIS problem in a differentiable manner by relaxing the embedding dimension search space to be continuous and optimizing the selection of dimensions with gradient descent.
%	\item 
%	to relax the NIS search space to be continuous, which allows searching for varying feature dimensions automatically in a differentiable manner with gradient descent.
%	\item We introduce a soft selection layer to optimize the selection of embedding dimensions for different features. A gradient normalization technique is proposed to keep the backpropagated gradients steady during the training of the soft selection layer. % and effective for optimizing the selection layer.
	\item We propose a gradient normalization technique to deal with the high variance of gradients during optimizing the soft selection layer, and further design a fine-grained pruning procedure through layer merging to produce a fine-grained mixed embedding dimension scheme for different features.
	\item The proposed method is model-agnostic, and thus can be incorporated with various existing architectures of latent factor models to improve recommendation performance and reduce the size of embedding parameters.
%	\item We conduct experiments with different model architectures on CF and CTR prediction tasks. The results demonstrate our \method method outperforms the existing NIS baselines in terms of recommendation performance, training efficiency and parameter size. 
	\item We conduct experiments with different model architectures on real-world datasets for three typical recommendation tasks: rating prediction, CTR prediction, and top-k item recommendation. The results demonstrate \method achieves the best overall result compared with existing NIS methods in terms of recommendation performance, embedding parameter size and training time cost.
%	\item 
%	compared with existing neural input search methods with fewer embedding parameters and less time cost.
%	achieves the best recommendation performance and highest computational efficiency with the least parameters compared with all the neural input search baselines.
\end{itemize}

\section{Differentiable Neural Input Search}
%\vspace{-0.05in}
%%\vspace
%\subsection{Latent Factor Models}
\subsection{Background}

\noindent{\bf{Latent factor models.}}
We consider a recommender system involving $M$ \emph{feature fields} (e.g., user ID, item ID, item price). 
Typically, $M$ is $2$ (including user ID and item ID) in collaborative filtering (CF) problems, whereas in the context of CTR prediction, $M$ is usually much larger than $2$ to include more feature fields. 
Each categorical feature field consists of a collection of discrete features, while a numerical feature field contains one scalar feature. Let ${\cal F}$ denote the list of features over all the fields and the size of ${\cal F}$ is $N$. 
% $x_{i}^{{j}}$ where $i$ is the feature id, and $j$ is its belong feature field id.??
For the $i$-th feature in ${\cal F}$, its initial representation is a $N$-dimensional sparse vector ${\bf x}_i$, where the $i$-th element is 1 (for discrete feature) or a scalar number (for scalar feature), and the others are 0s.
Latent factor models generally consists of two parts: one feature embedding layer, followed by the feature interaction layers. Without loss of generality, the input instances to the latent factor model include several features belonging to the respective feature fields. The feature embedding layer transforms all the features in an input instance into dense embedding vectors. Specifically, a sparsely encoded input feature vector ${\bf x}_i\in \mathbb{R}^N$ is transformed into a $K$-dimensional embedding vector ${\bf e}_{i}\in \mathbb{R}^{K}$ as follows:
\begin{equation}\label{equ:normal_embedding}
%\small
{\bf e}_{i} = {\bf E}^\top {\bf x}_{i}
\end{equation}
where ${\bf E}\in \mathbb{R}^{N\times K}$ is known as the \emph{embedding matrix}. %{\bf e}_{i_m}^{(j)}
The output of the feature embedding layer is the collection of dense embedding vectors for all the input features, which is denoted as ${\bf X}$.
%\begin{equation}\label{equ:embed_output}
%\mathbf{e} = \{\mathbf{e}^{(1)},\mathbf{e}^{(2)},...,\mathbf{e}^{(m)}\}
%\end{equation}
%where $\mathbf{e}^{(j)}\in \mathbb{R}^{K}$ is the embedding of the input feature in the $j$-th field and $j\in[1,m]$.
%
The feature interaction layers, which are designed to be different architectures, essentially compose a parameterized function $\cal{G}$ that predicts the objective based on the collected dense feature embeddings ${\bf X}$ for the input instance. That is, 
%$f$ that predicts the objective given collected feature embeddings:
\begin{equation}\label{equ:prediction}
%\small
\hat{y} = \cal{G}(\bm{\theta}, \mathbf{X})
\end{equation}
where $\hat{y}$ is the model's prediction, and $\bm{\theta}$ denotes the set of parameters in the interaction layers. Prior works have developed various architectures for $\cal{G}$, including the simple inner product function~\cite{FM}, and deep neural networks-based interaction  functions~\cite{DBLP:conf/www/HeLZNHC17,DBLP:conf/ijcai/ChengSZH18,xDeepFM,DBLP:conf/recsys/Cheng0HSCAACCIA16,DBLP:conf/ijcai/GuoTYLH17}. 
Most of the proposed architectures for the interaction layers require all the feature embeddings to be in a uniform dimension.

\noindent{\bf{Neural architecture search.}} Neural Architecture Search (NAS) has been proposed to automatically search for the best neural network architecture. 
To explore the space of neural architectures, different search strategies have been explored including random search~\cite{DBLP:conf/uai/LiT19}, evolutionary methods~\cite{DBLP:conf/iclr/ElskenMH19,DBLP:conf/icga/MillerTH89,DBLP:conf/icml/RealMSSSTLK17}, Bayesian optimization~\cite{DBLP:conf/icml/BergstraYC13,DBLP:conf/ijcai/DomhanSH15,DBLP:conf/icml/MendozaKFSH16}, reinforcement learning~\cite{DBLP:conf/iclr/BakerGNR17,DBLP:conf/cvpr/ZhongYWSL18,DBLP:conf/iclr/ZophL17}, and gradient-based methods~\cite{DBLP:conf/iclr/CaiZH19,DBLP:conf/iclr/LiuSY19,DBLP:conf/iclr/XieZLL19}. Since being proposed in~\cite{DBLP:conf/iclr/BakerGNR17,DBLP:conf/iclr/ZophL17}, NAS has achieved remarkable performance in various tasks such as image classification~\cite{DBLP:conf/aaai/RealAHL19,DBLP:conf/cvpr/ZophVSL18}, semantic segmentation~\cite{DBLP:conf/nips/ChenCZPZSAS18} and object detection~\cite{DBLP:conf/cvpr/ZophVSL18}.
However, most of these researches have focused on searching for optimal network structures automatically, while little attention has been paid to the design of the input component. This is because the input component in visual tasks is already given in the form of floating point values of image pixels. As for recommender systems, an input component based on the embedding layer is deliberately developed to transform raw features (e.g., discrete user identifiers) into dense embeddings. In this paper, we focus on the problem of neural input search, which can be considered as NAS on the input component (i.e., the embedding layer) of recommender systems.

\subsection{Search Space and Problem}
%\vspace{-0.05in}  

\begin{figure}
	\centering
	\subcaptionbox{Example of notations.\label{fig:model:a}}
	[.38\linewidth]{\includegraphics[width=.38\linewidth]{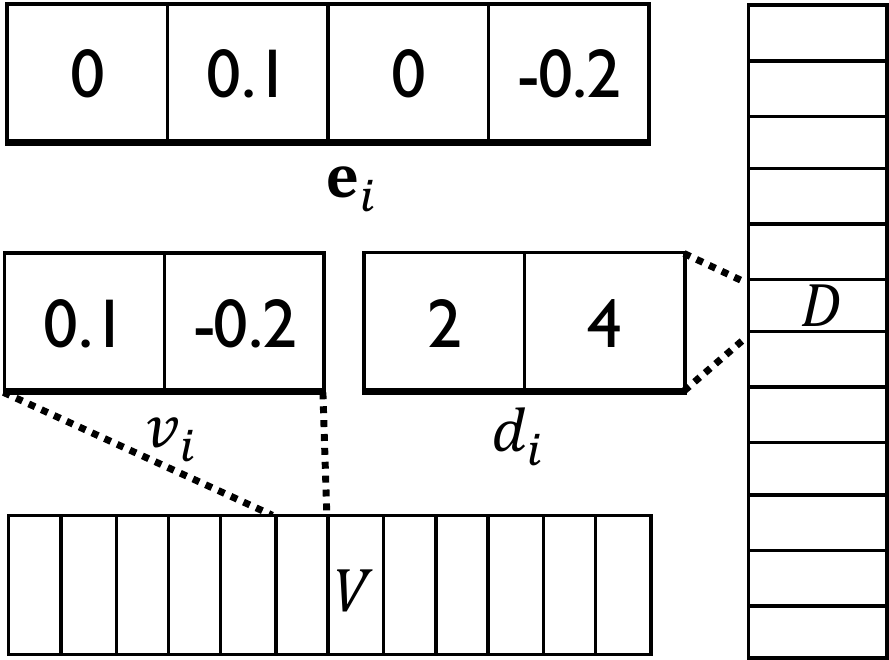}}
%	\quad
	\subcaptionbox{Model structure.\label{fig:model:b}}
	[.61\linewidth]{\includegraphics[width=.61\linewidth]{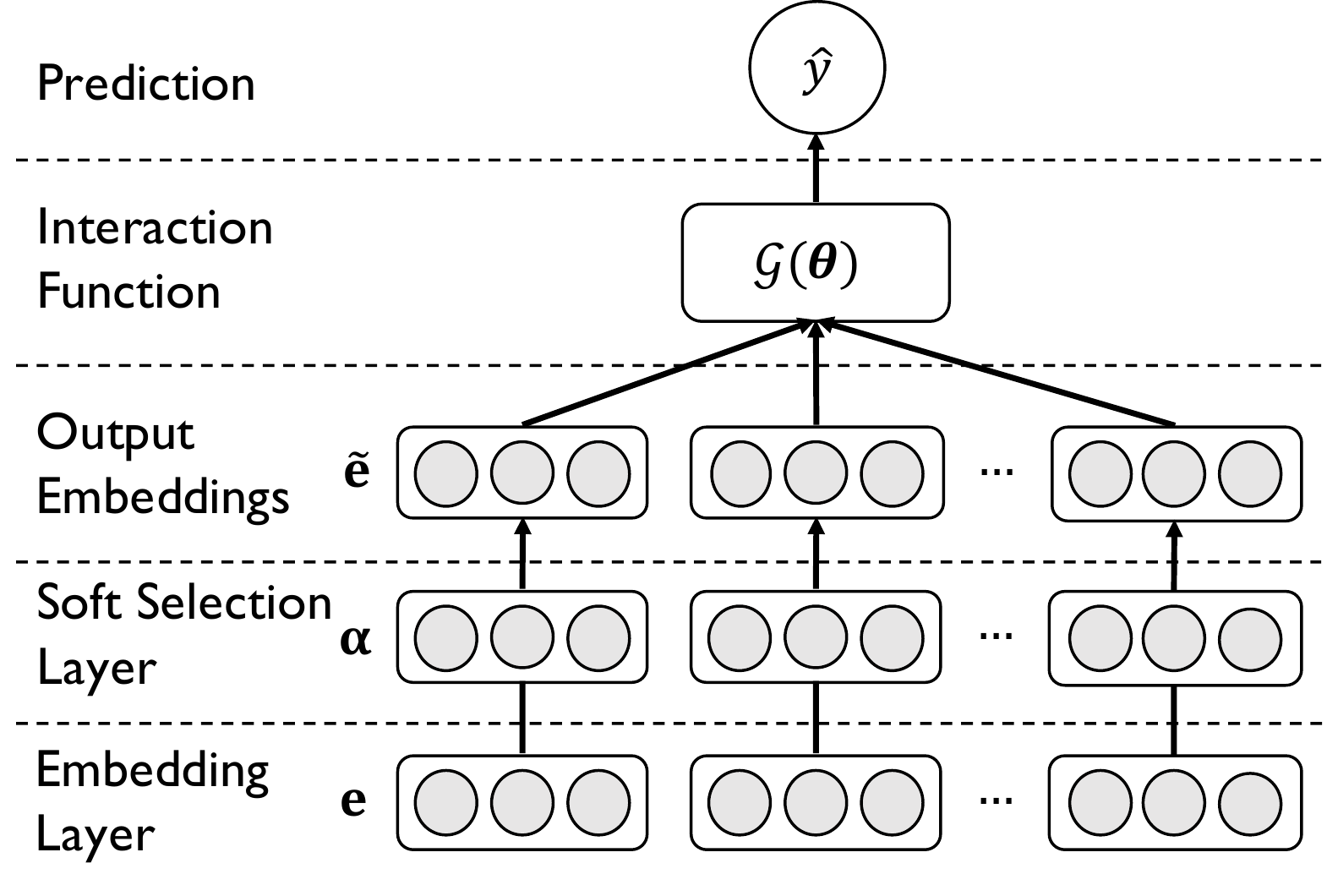}}
		%\vspace{-0.05in}
	\caption{A demonstration of notations and model structure.}\label{fig:model}
\vspace{-0.1in}
\end{figure}

\noindent{\bf{Search space.}}
The key idea of neural input search is to use embeddings with mixed dimensions to represent different features.
To formulate feature embeddings with different dimensions, we adopt the representation for sparse vectors 
%we represent them as sparse representations of uniform dimension vectors 
(with a \emph{base dimension} $K$).
%, which can be absorbed by existing feature interaction layers.
%shenyy:check
Specifically, for each feature, we maintain a \emph{dimension index vector} $d$ which contains ordered locations of the feature's existing dimensions from the set $\{1,\cdots,K\}$, and an \emph{embedding value vector} $v$ which stores embedding values in the respective existing dimensions.
% and these varying-length embeddings are essentially the sparse representations of equal-length vectors (with a \emph{base dimension} $K$) absorbed by the feature interaction layers. 
%A common practice to represent feature embeddings with different dimensions is to maintain an \emph{index vector} and a \emph{value vector} per feature. 
% Specifically, for any feature, its index vector contains ordered locations of the existing dimensions from the set of $\{1,\cdots,K\}$, and its value vector stores the embedding values in the respective existing dimensions. 
% The number of existing dimensions varies among different features. 
The conversion from the index and value vectors of a feature into the $K$-dimensional embedding vector $\mathbf{e}$ is straightforward. Figure~\ref{fig:model:a} gives an example of $d_i$, $v_i$ and $\mathbf{e}_i$ for the $i$-th feature in $\cal{F}$. Note that ${\bf e}$ corresponds to a row in the embedding matrix ${\bf E}$.

% \begin{figure}[t!]
%	\centering{\includegraphics[width=.32\linewidth]{figures/notation.pdf}}
%	\caption{An example of problem notations.}\label{fig:model:a}
%	%\vspace{-0.2in}
%\end{figure} 
%  
% \begin{figure}[t!]
%	\centering{\includegraphics[width=.9\linewidth]{figures/model.pdf}}	\caption{Model Structure.}\label{fig:model:b}
%	%\vspace{-0.2in}
%\end{figure} 

%  It is also worth noticing that the space cost of index and value vectors can be low when most feature embeddings only involve small numbers of existing dimensions. 

%In order to simplify the definition of the search space in neural input search, we transform the index and value vectors of any feature into two respective $K$-dimensional vectors ${\bf d}$ and ${\bf v}$. ${\bf d}\in \{0,1\}^{K}$ is a binary vector indicating the existence of each embedding dimension, and ${\bf v}\in\mathbb{R}^{K}$ is a numerical vector storing all the embedding values where the values for non-existent dimensions are 0s. 
%Figure~\ref{fig:model:a} shows an example of ${\bf d}$ and ${\bf v}$ of a feature. 
%Henceforth, we refer to ${\bf d}_i$ and ${\bf v}_i$ as the \emph{dimension index vector} and \emph{dimension value vector} for the $i$-th feature in $\cal{F}$, respectively. 

The size of $d$ varies among different features to enforce a mixed dimension scheme. Formally, given the feature set ${\cal F}$, we define the \emph{mixed dimension scheme} $D=\{d_1, \cdots, d_N\}$ to be the collection of dimension index vectors for all the features in ${\cal F}$.
% which can be organized into a matrix ${\bf D}\in\{0,1\}^{N\times K}$. 
We use $\mathcal{D}$ to denote the \emph{search space} of the mixed dimension scheme $D$ for ${\cal F}$, which includes $2^{N K}$ possible choices. Besides, we denote by $V=\{v_1, \cdots, v_N\}$ the set of the embedding value vectors for all the features in ${\cal F}$. Then we can derive the embedding matrix $\mathbf{E}$ with $D$ and $V$ to make use of conventional feature interaction layers.
 
\noindent{\bf{Problem formulation.}}
Let $\Theta=\{\bm{\theta}, V\}$ be the set of trainable model parameters, and $\mathcal{L}_{train}$ and $\mathcal{L}_{val}$ are model's training loss and validation loss, respectively. The two losses are determined by both the mixed dimension scheme $D$, and the trainable parameters $\Theta$.
The goal of neural input search is to find a mixed dimension scheme $D \in \mathcal{D}$ that minimizes the validation loss $\mathcal{L}_{val} (\Theta^*,D)$, where the parameters $\Theta^*$ given any mixed dimension scheme are obtained by minimizing the training loss.  
%$\Theta^* = \mathop{\arg\!\min}_{\Theta}\mathcal{L}_{train}(\Theta,D)$.
This can be formulated as:
%
%The training of a latent factor model is to learn the parameters $\bm\theta$ in the interaction layers, and the embedding matrix ${\bf E}$. 
%%Recall that we use $\cal{G}(\bm{\theta}, {\bf E})$ to denote the interaction function, where $\bm{\theta}$ contains all the parameters in the interaction layers of a latent factor model. 
%Given the mixed dimension scheme ${\bf D}$, let $\Theta=\{\bm{\theta},\mathbf{E}\}$ be the set of trainable model parameters, and $\mathcal{L}_{train}$ and $\mathcal{L}_{val}$ are model's training loss and validation loss, respectively. The problem of this paper is to find a mixed dimension scheme $D$ in the search space $\mathcal{D}$ to minimize model's loss on the validation set:
\begin{equation}\label{equ:definition}
%\small
\begin{aligned}
	\min_{D \in \mathcal{D}}\ \ \ &\mathcal{L}_{val} (\Theta^*(D),D)\\
	\text{s.t.}\ \ \ &\Theta^*(D) = \mathop{\arg\!\min}_{\Theta}\mathcal{L}_{train}(\Theta,D) 
\end{aligned} 
\end{equation}
% Consider $\mathcal{L}_{train}$ and $\mathcal{L}_{val}$ to be model's loss on training set and validation set, respectively. Let the interaction function be $f(\bm{\theta})$, where $\bm{\theta}$ are parameters in the interaction layers, and let $\Theta=\{\bm{\theta},\tilde{\mathbf{e}} = \{\tilde{\mathbf{e}}_1,\tilde{\mathbf{e}}_2,...,\tilde{\mathbf{e}}_v\}\}$ to be the set of trainable parameters in the model, we aim to find mixed dimension scheme $\mathcal{D}=\{d_1,d_2,...,d_v\}$ and corresponding projection matrices $\mathcal{P}=\{\mathbf{P}_1, \mathbf{P}_2,...,\mathbf{P}_v\}$ to minimize model's loss on the validation set:
%\begin{equation}\label{equ:definition}
%\begin{aligned}
%	\min_{\mathcal{D}, \mathcal{P}}\ \ \ &\mathcal{L}_{val}(\Theta^*(\mathcal{D}, \mathcal{P}),\mathcal{D}, \mathcal{P})\\
%	\text{s.t.}\ \ \ &\Theta^*(\mathcal{D}, \mathcal{P}) = \mathop{\arg\!\min}_{\Theta}\mathcal{L}_{train}(\Theta,\mathcal{D}, \mathcal{P}) 
%\end{aligned}
%\end{equation}
%Given the set of input features $\mathcal{F}$ Find a scheme of mixed dimension scheme embeddings
The above problem formulation is actually consistent with hyperparameter optimization in a broader scope~\cite{DBLP:conf/icml/MaclaurinDA15,DBLP:conf/icml/Pedregosa16,DBLP:conf/icml/FranceschiFSGP18}, since the mixed dimension scheme $D$ can be considered as model hyperparameters to be determined according to model's validation performance. However, the main difference is that the search space $\mathcal{D}$ in our problem is much larger than the search space of conventional hyperparameter optimization problems.
\subsection{Feature Blocking}
%\vspace{-0.05in}

Feature blocking has been a novel ingredient used in the existing neural input search methods~\cite{DBLP:conf/kdd/JoglekarLCXWAKL20,ginart2019mixed} to facilitate the reduction of search space. The intuition behind is that features with similar frequencies could be grouped into a block sharing the same embedding dimension. %number of embedding dimension. 
Following the existing works, we first employ feature blocking to control the search space of the mixed dimension scheme. We sort all the features in ${\cal F}$ in the descending order of frequency (i.e., the number of feature occurrences in the training instances). Let $\eta_f$ denote the frequency of feature $f\in\cal{F}$. 
We can obtain a sorted list of features $\tilde{\cal F} = [f_1, f_2, \cdots, f_N]$ such that $\eta_{f_i}\geq \eta_{f_j}$ for any $i<j$. 
We then separate $\tilde{\cal F}$ into $L$ blocks, where the features in a block  share the same dimension index vector $d$. We denote by $\tilde{D}$ the mixed dimension scheme after feature blocking. Then the length of the mixed dimension scheme $|\tilde{D}|$ becomes $L$, and the search space size is reduced from $2^{NK}$ to $2^{LK}$ accordingly, where $L\ll N$.

\eat{
In the first step of our approach to DNIS, we employ feature blocking to reduce the search space of the mixed dimension scheme.
Feature blocking has been a common practice in existing works of neural input search~\cite{DBLP:conf/kdd/JoglekarLCXWAKL20,ginart2019mixed}, considering that features with similar frequency can be represented using similar embedding dimensions.
We first sort and reorder features according to their frequency (i.e., times of appearance in the input of training data) in decreasing order to get $\mathcal{F}_{sorted}=[\mathbf{x}_1,\mathbf{x}_2,...,\mathbf{x}_v]$, where $\mathbf{f}_{\mathbf{x}_1}>=\mathbf{f}_{\mathbf{x}_2}>=...>=\mathbf{f}_{\mathbf{x}_v}$, and $\mathbf{f}_{\mathbf{x}_i}$ denotes the frequency of $\mathbf{x}_i$. 
Then we divide the sorted features into multiple blocks, where features in each block will share an embedding dimension index $D$. 
Recall that $D\in\mathbb{R}^{n\times K}$,
suppose we divide features into $k$ blocks, then 
the mixed dimension scheme after blocking is changed to $\tilde{D}\in\mathbb{R}^{k\times K}$.
%the size of $\mathcal{D}$ can be reduced from $2^{vK}$ to $2^{kK}$. 
%Note that, although here we share the dimension index $D$ in each block during training, \method supports non-uniform dimension index inside each block through parameter merging and pruning, which will be further clarified in Section~\ref{sec:param_merge}.
}
 %\vspace{-0.05in}
\subsection{Continuous Relaxation and Differentiable Optimization}

\textbf{Continuous relaxation.} 
After feature blocking, in order to optimize the mixed dimension scheme $\tilde{D}$, we first transform $\tilde{D}$ into a binary \emph{dimension indicator matrix} $\tilde{\mathbf{D}}\in \mathbb{R}^{L\times K}$,  where each element in $\tilde{\mathbf{D}}$ is either 1 or 0 indicating the existence of the corresponding embedding dimension according to $\tilde{D}$. We then introduce a \emph{soft selection layer} to relax the search space of $\tilde{\mathbf{D}}$ to be continuous.
% we introduce a \emph{soft selection layer} to relax the search space of $\tilde{D}$ to be continuous. First, we The soft selection layer is essentially a numerical matrix $\bm{\alpha}\in \mathbb{R}^{K\times K}$
%Recall that the dimension index vectors in $\tilde{D}$ include binary choices of embedding dimensions for the feature blocks. In order to optimize $\tilde{D}$, we introduce a \emph{soft selection layer} to relax the search space of $\tilde{D}$ to be continuous. 
The soft selection layer is essentially a numerical matrix $\bm{\alpha}\in \mathbb{R}^{L\times K}$, where each element in $\bm\alpha$ satisfies:
$0\leq {\bm \alpha}_{l,k}\leq 1$.
That is, each binary choice $\tilde{\mathbf{D}}_{l,k}$ (the existence of the $k$-th embedding dimension in the $l$-th feature block) in $\tilde{\mathbf{D}}$ is relaxed to be a continuous variable $\bm\alpha_{l,k}$ within the range of $[0,1]$. %$0$ and $1$.
%Next, we convert all the embedding value vectors in $V$ to a numerical \emph{embedding value matrix} ${\bf V}\in\mathbb{R}^{N\times K}$ according to $alpha$.
We insert the soft selection layer between the feature embedding layer and interaction layers in the latent factor model, as illustrated in Figure~\ref{fig:model:b}.
Given $\bm\alpha$ and the embedding matrix ${\bf E}$, the output embedding $\tilde{{\bf e}}_i$ of a feature $f_i$ in the $l$-th block produced by the bottom two layers can be computed as follows:
\begin{equation}\label{equ:soft_select}
\begin{aligned}
\tilde{\mathbf{e}}_i = {\mathbf{e}}_i\odot \bm{\alpha}_{l*}\\
%		&\text{s.t.}\ \ \tilde{\mathbf{e}}_i \in \tilde{\mathbf{E}}^{(j)}
\end{aligned}
\end{equation}
%%
%\eat{
%Besides, we set all the embedding value vectors in $V$ to full dimension and combine them into a primary embedding matrix $\tilde{\mathbf{E}}\in \mathbb{R}^{N\times K}$. Recall that we can construct the embedding matrix $\mathbf{E}$ with mixed dimension scheme $D$ and the set of value vectors $V$, now we can similarly produce $\mathbf{E}$ with $\bm\alpha$ and $\tilde{\mathbf{E}}$. Formally, 
%%%\in[0,1]. 
%%We insert the softAaa selection layer between the feature embedding layer and interaction layers in the latent factor model, as shown in Figure~\ref{fig:model:b}. By combining $\bm\alpha$  with the embedding matrix ${\bf E}$, 
%the output embedding of a feature $f_i$ in the $l$-th block is the following:
%\begin{equation}\label{equ:soft_select}
%\small
%\begin{aligned}
%\mathbf{e}_i = \tilde{\mathbf{e}}_i\odot \bm{\alpha}_{l*}\\
%%		&\text{s.t.}\ \ \tilde{\mathbf{e}}_i \in \tilde{\mathbf{E}}^{(j)}
%\end{aligned}
%\end{equation}
%}
where $\bm{\alpha}_{l*}$ is the $l$-th row in $\bm{\alpha}$, and $\odot$ is the element-wise product.
By applying Equation (\ref{equ:soft_select}) to all the input features, we can obtain the output feature embeddings ${\bf X}$.
%After the soft selection layer, we are able to obtain the embeddings for all the input features, which compose ${\bf X}$. 
Next, we supply ${\bf X}$ to the feature interaction layers for final prediction as specified in Equation~(\ref{equ:prediction}).
Note that $\bm\alpha$ is used to softly select the dimensions of feature embeddings during model training, and the discrete mixed dimension scheme $D$ will be derived after training.

\noindent{\bf{Differentiable optimization.}} 
Now that we relax the mixed dimension scheme $\tilde{\bf D}$ (after feature blocking) via the soft selection layer $\bm \alpha$, %, and transform the embedding value vectors in $V$ to the primary embedding matrix $\tilde{\mathbf{E}}$.
our problem stated in Equation~(\ref{equ:definition}) can be transformed into:
\begin{equation}\label{equ:problem_transform}
%\small
\begin{aligned}
\min_{\bm{\alpha}}\ \ \ &\mathcal{L}_{val}(\Theta^*(\bm{\alpha}),\bm{\alpha})\\
\text{s.t.}\ \ \ &\Theta^*(\bm{\alpha}) = \mathop{\arg\!\min}_{\Theta}\mathcal{L}_{train}(\Theta,\bm{\alpha}) \wedge \bm{\alpha}_{k,j}\in[0,1]
\end{aligned} 
\end{equation} 
%%
%\eat{
%Note that after the feature blocking and continuous relaxation, we use the soft selection layer $\bm{\alpha}$ to encode the mixed dimension scheme $D$. According to Equation~(\ref{equ:definition}), our problem is transformed into:
%\begin{equation}\label{equ:problem_transform}
%\begin{aligned}
%	\min_{\bm{\alpha}}\ \ \ &\mathcal{L}_{val}(\Theta^*(\bm{\alpha}),\bm{\alpha})\\
%	\text{s.t.}\ \ \ &\Theta^*(\bm{\alpha}) = \mathop{\arg\!\min}_{\Theta}\mathcal{L}_{train}(\Theta,\bm{\alpha}), 0\leq\bm{\alpha}_{i,j}\leq 1
%\end{aligned} 
%\end{equation} 
%}
where $\Theta=\{\bm{\theta},{\mathbf{E}}\}$ represents model parameters in both the embedding layer and interaction layers. Equation~\ref{equ:problem_transform} essentially defines a bi-level optimization problem~\cite{DBLP:journals/anor/ColsonMS07}, which has been studied in differentiable NAS~\cite{DBLP:conf/iclr/LiuSY19} and gradient-based hyperparameter optimization~\cite{DBLP:conf/kdd/ChenC0GLLW19,DBLP:conf/icml/FranceschiFSGP18,DBLP:conf/icml/Pedregosa16}.
%Algorithm~\ref{alg} summarizes the alternating optimization procedure for solving Equation (\ref{equ:problem_transform}).
Basically, $\bm\alpha$ and $\Theta$ are respectively treated as the upper-level and lower-level variables to be optimized in an interleaving way. 
To deal with the expensive optimization of $\Theta$, we follow the common practice that approximates $\Theta^*(\bm{\alpha})$ by adapting $\Theta$ using a single training step:
%and employ an alternating optimization. Formally:
\begin{equation}
%\small
	\Theta^*(\bm{\alpha}) \approx \Theta-\xi \nabla_{\Theta}\mathcal{L}_{train}(\Theta,\bm{\alpha})
\end{equation}
where $\xi$ is the learning rate for one-step update of model parameters $\Theta$. Then we can optimize $\bm{\alpha}$ based on the following gradient: %by descending:
\begin{equation}\label{equ:gradient}
%\small
\begin{aligned} 
		&\nabla_{\bm{\alpha}}\mathcal{L}_{val}(\Theta-\xi \nabla_{\Theta}\mathcal{L}_{train}(\Theta,\bm{\alpha}),\bm{\alpha})\\
		=&\nabla_{\bm{\alpha}}\mathcal{L}_{val}(\Theta',\bm{\alpha})-\xi\nabla^2_{\bm{\alpha},\Theta}\mathcal{L}_{train}(\Theta,\bm{\alpha})\cdot \nabla_{\bm{\alpha}}\mathcal{L}_{val}(\Theta',\bm{\alpha})
\end{aligned}
\end{equation}
where $\Theta'=\Theta-\xi \nabla_{\Theta}\mathcal{L}_{train}(\Theta,\bm{\alpha})$ denotes the model parameters after one-step update. Equation (\ref{equ:gradient}) can be solved efficiently using the existing deep learning libraries that allow automatic differentiation, such as Pytorch~\cite{DBLP:conf/nips/PaszkeGMLBCKLGA19}. The second-order derivative term in Equation (\ref{equ:gradient}) can be omitted to further improve computational efficiency considering $\xi$ to be near zero, which is called the \emph{first-order approximation}.
% In this paper, we adopt the first-order approximation in \method by default since we find the final performance is similar with and without the approximation.
Algorithm~\ref{alg} (line 5-10) summarizes the bi-level optimization procedure for solving Equation (\ref{equ:problem_transform}).
 
\noindent{\bf{\emph{Gradient normalization.}}} During the optimization of $\bm{\alpha}$ by the gradient $\nabla_{\bm{\alpha}}\mathcal{L}_{val}(\Theta',\bm{\alpha})$, we propose a gradient normalization technique to normalize the row-wise gradients of $\bm{\alpha}$ over each training batch:
\begin{equation}\label{eq:norm}
%\small
	{\bf g}_{norm}(\bm{\alpha}_{l*}) = \frac{{\bf g}(\bm{\alpha}_{l*})}{\Sigma^{K}_{k=1} |{\bf g}(\bm{\alpha}_{l,k})| /K+\epsilon_{\bf g}}, ~~~~k\in[1,K]
\end{equation}
where ${\bf g}$ and ${\bf g}_{norm}$ denote the gradients before and after normalization respectively, and 
%recall $\bm{\alpha}_{j*}$ is the $j$-th row of $\bm{\alpha}$, which corresponds to the $j$-th feature block, and
$\epsilon_{\bf g}$ is a small value (e.g., 1e-7) to avoid numerical overflow. Here we use row-wise gradient normalization to deal with the high variance of the gradients of $\bm{\alpha}$ during backpropogation. More specifically, $g(\bm{\alpha}_{l*})$ of a high-frequency feature block can be several orders of magnitude larger than that of a low-frequency feature block due to their difference on the number of related data samples.
% is high due to the significant difference in feature frequency.
%The consideration is that the gradients of $\bm{\alpha}_{l*}$ varies a lot over different feature blocks due to the significant difference in feature frequency. 
By normalizing the gradients for each block, we can apply the same learning rate to different rows of $\bm{\alpha}$ during optimization. Otherwise, a single learning rate shared by different feature blocks will fail to optimize most rows of $\bm{\alpha}$.
% may easily fall short in optimizing all the rows of $\bm{\alpha}$.
%Otherwise, using a same learning rate for different feature blocks will fail to optimize most rows of $\bm{\alpha}$.
%Otherwise, we have to tune fine-grained learning rates for different feature blocks to optimize the soft selection layer.
\newlength{\textfloatsepsave}
\setlength{\textfloatsepsave}{\textfloatsep}
\setlength{\textfloatsep}{0pt}
\begin{algorithm}[t]\label{alg:optimize}
%\small
	\SetAlgoLined
	\SetNlSty{}{}{:}
	\ShowLn\KwIn{training dataset, validation dataset.}
	\ShowLn\KwOut{mixed dimension scheme $D$, embedding values $V$, interaction function parameters $\mathbf{\theta}$.}
	%Sort and reorder features according to popularity, get $\mathcal{F}_{sorted}$\;
	%Divide features $\mathcal{F}_{sorted}$ into $k$ blocks\;
	Sort features into $\tilde{\cal F}$ and divide them into $L$ blocks\;
	Initialize the soft selection layer $\bm{\alpha}$ to be an all-one matrix, and randomly initialize $\Theta$\tcp*[l]{$\Theta=\{\bm{\theta},{\mathbf{E}}\}$}
	\While{not converged}{
		Update trainable parameters $\Theta$ by descending $\nabla_{\Theta}\mathcal{L}_{train}(\Theta,\bm{\alpha})$\;
		Calculate the gradients of $\bm{\alpha}$ as: $-\xi\nabla^2_{\bm{\alpha},\Theta}\mathcal{L}_{train}(\Theta,\bm{\alpha})\cdot \nabla_{\bm{\alpha}}\mathcal{L}_{val}(\Theta',\bm{\alpha})+\nabla_{\bm{\alpha}}\mathcal{L}_{val}(\Theta',\bm{\alpha})$\;
		%$\nabla_{\bm{\alpha}}\mathcal{L}_{val}(\Theta^*(\bm{\alpha}),\bm{\alpha})$
		%  \Indp{($\Theta^*(\bm{\alpha})=\Theta$ if using first-order approximation)}
		\tcp{(set $\xi=0$ if using first-order approximation)} 
	%	Clip $\mathbf{\alpha}$ into the range of $[0,1]$\;
		Perform Equation~(\ref{eq:norm}) to normalize the gradients in ${\bm\alpha}$\;
		Update $\bm{\alpha}$ by descending the gradients, and then clip its values into the range of $[0,1]$\;
		
	} 
	Calculate the output embedding matrix $\mathbf{E}$ using $\bm{\alpha}$ and $\tilde{\mathbf{E}}$ according to Equation (\ref{equ:soft_select})\;
	Prune $\mathbf{E}$ into a sparse matrix $\mathbf{E'}$ following Equation (\ref{equ:pruneindex})\; 
	Derive the mixed dimension scheme $D$ and embedding values $V$ with $\mathbf{E'}$\; 
%	\KwRet{$D$, $V$, $\bm{\theta}$}
	\caption{DNIS - Differentiable Neural Input Search}\label{alg}
\end{algorithm}

\subsection{Deriving Fine-grained Mixed Embedding Dimension Scheme}\label{sec:param_merge}
%\vspace{-0.12in}
%\subsection{Deriving Feature Embeddings in Mixed Dimensions}\label{sec:param_merge}
%\subsection{Parameter Merging, Pruning and Saving}\label{sec:param_merge}
%\vspace{-0.05in}
After optimization, we have the learned parameters for $\bm\theta$, ${{\bf E}}$ and $\bm{\alpha}$. 
A straightforward way to derive the discrete mixed dimension scheme ${D}$ is to prune non-informative embedding dimensions in the soft selection layer $\bm{\alpha}$. Here we employ a fine-grained pruning procedure through layer merging.
%This allows us to derive the discrete mixed dimension scheme ${D}$. % for all the features in ${\cal F}$.
Specifically, for feature $f_i$ in the $l$-th block, we can compute its output embedding $\tilde{\mathbf{e}}_i$ with ${\mathbf{e}}_i$ and $\bm{\alpha}_{l*}$ following Equation (\ref{equ:soft_select}). 
%By merging the embedding layer with the soft selection layer, 
We collect the output embeddings $\tilde{\mathbf{e}}_i$ for all the features in ${\bf \cal{F}}$ and form an output embedding matrix $\tilde{\bf E}\in\mathbb{R}^{N\times K}$. We then prune non-informative embedding dimensions in $\tilde{\bf E}$ as follows:
% for all the features in ${\cal F}$ into an output embedding matrix $\mathbf{E}$. Then we prune non-informative embedding dimensions in $\mathbf{E}$ following:
\begin{equation}\label{equ:pruneindex}
%\small
\tilde{\mathbf{E}}_{i,j}=
\begin{cases}  
0, & {\rm if~}\ |\tilde{\mathbf{E}}_{i,j}|<\epsilon \\
\tilde{\mathbf{E}}_{i,j}, & {\rm otherwise} 
\end{cases}
\end{equation}
where $\epsilon$ is a threshold that can be manually tuned according to the requirements on model performance and parameter size.
The pruned output embedding matrix $\tilde{\bf E}$ is sparse and can be used to derive the discrete mixed dimension scheme $D$ and the embedding value vectors $V$ for $\cal{F}$ accordingly.
With fine-grained pruning, the derived embedding dimensions can be different even for features in the same feature block, resulting in a more flexible mixed dimension scheme.

\noindent{\bf{Relation to network pruning.}} Network pruning, as one kind of model compression techniques, improves the efficiency of over-parameterized deep neural networks by removing redundant neurons or connections without damaging model performance~\cite{cheng2017survey,DBLP:conf/iclr/LiuSZHD19,DBLP:conf/iclr/FrankleC19}.
% Some researches~\cite{DBLP:conf/iclr/LiuSZHD19,DBLP:conf/iclr/FrankleC19} indicate that network pruning can be also considered as performing implicit architecture search. 
%Early studies on network pruning~\cite{DBLP:conf/nips/HansonP88,DBLP:conf/nips/CunDS89,DBLP:conf/nips/HassibiS92} used weight decay or Hessian of loss function to select and remove redundant connections during training. 
Recent works of network pruning~\cite{han2015learning,DBLP:conf/iclr/MolchanovTKAK17,DBLP:conf/iclr/0022KDSG17} generally performed iterative pruning and finetuning over certain pretrained over-parameterized deep network. 
%Most existing network pruning methods require manual configurations of pruning thresholds and l1 or l2 regularization terms for layers. 
%\method optimizes feature embeddings with the gradients from the validation set, which benefits the selection of predictive dimensions, instead of simply removing redundant weights in the embeddings.
Instead of simply removing redundant weights,
our proposed method \method optimizes feature embeddings with the gradients from the validation set, and only prunes non-informative embedding dimensions and their values in one shot after model training. 
This also avoids manually tuning thresholds and regularization terms per iteration.
We conduct experiments to compare the performance of \method and network pruning methods in Section~\ref{section/compare_prune}.
%Furthermore, our DNIS can be seamlessly incorporated with various latent factor models (with different architectures for feature interaction layers). 

\eat{
 which also prunes non-informative weights, only prune the redundant embedding values in one shot after training and embedding merging, avoiding iterative pruning and manual tuning of pruning sensitivity (e.g., pruning thresholds and regularization terms in each iteration). 
% which focuses on designing the embedding layer, prunes non-informative values in dense embeddings after training and embedding merging in one shot, avoiding manual tuning of pruning sensitivity (i.e., pruning thresholds and regularization terms) and iterative pruning. 
We have also provided an experimental comparison of \method with network pruning in Section~\ref{section/compare_prune}.
}

%\vspace{-0.1in}
\section{Experiments}
%%\vspace{-0.125in} 
%
%In this section, we conduct experiments to answer the following research questions:\\
%\noindent{\bf RQ1:} How does our method \method perform against the existing methods?\\
%\noindent{\bf RQ2:} How does the performance of \method vary with different settings of hyperparameters?\\
%\noindent{\bf RQ3:} What are the learned feature dimensions, and how do they differ from network pruning results?
\subsection{Experimental Settings}
%\vspace{-0.05in}
\noindent{\bf Datasets.} 
We used two benchmark datasets Movielens-20M~\cite{DBLP:journals/tiis/HarperK16} and Criteo~\cite{criteo} 
for rating prediction and CTR prediction tasks, respectively. 
For each dataset, we randomly split the instances by 8:1:1 to obtain the training, validation and test sets. Besides, we also conduct experiments on Movielens-1M dataset~\cite{DBLP:journals/tiis/HarperK16} to compare with NIS-ME~\cite{DBLP:conf/kdd/JoglekarLCXWAKL20} for top-k item recommendation task. The statistics of the three datasets are summarized in Table~\ref{tab:datasets}.\\
{(1) \bf Movielens-20M} is a CF dataset containing more than 20 million user ratings ranging from 1 to 5  on movies.\\
%This is the 20 million version of Movielens datasets. The dataset consists of users' ratings, ranging from 1 to 5, on different movies.\\
{(2) \bf Criteo} is a popular industry benchmark dataset for CTR prediction, which contains 13 numerical feature fields and 26 categorical feature fields. Each label indicates whether a user has clicked the corresponding item.\\
{(3) \bf Movielens-1M} is a CF dataset containing over one million user ratings ranging from 1 to 5 on movies.
%, we follow~\cite{DBLP:conf/kdd/JoglekarLCXWAKL20} to process it into an implicit feedback dataset.

\setlength{\textfloatsep}{16pt}
\begin{table}[t]
%\small
	\captionsetup{labelfont=bf}
	\centering
%		%\vspace{-0.15in}
		\caption{Statistics of the datasets.}
			\label{tab:datasets}
			\resizebox{.8\linewidth}{!}{
	\begin{tabular}{cccc}
		\toprule[1pt]
		Dataset&\!Instance\#\!&\!Field\#\!&\!Feature\#\!\\
		\midrule[0.5pt] 
		Movielens-20M&20,000,263&2&165,771\\
		Criteo&45,840,617&39&2,086,936\\
			Movielens-1M&1,000,209&2&9,746\\
		\bottomrule[1pt]
	\end{tabular}
	}
\vspace{-0.1in}
\end{table}

\noindent{\bf{Evaluation metrics.}}
We adopt MSE (mean squared error) for rating prediction task, and use AUC (Area Under the ROC Curve) and Logloss for CTR prediction task. In addition to predictive performance, we also report the embedding parameter size and the overall time cost of each method. When comparing with NIS-ME, we provide Recall, MRR (mean reciprocal rank) and NDCG results for top-k recommendation.

\noindent{\bf Comparison methods.}
We compare our \method method with the following four approaches.

%, which are briefly described as follows. \\ 
{$\bullet$ \bf Grid Search}. This is the traditional approach to searching for a uniform embedding dimension. In our experiments, we searched 16 groups of dimensions, ranging from 4 to 64 with a stride of 4.

%我们实验中从4到64维，以4为步长搜索了16组参数
{$\bullet$ \bf Random Search}. Random search has been recognized as a strong baseline for NAS problems~\cite{DBLP:conf/iclr/LiuSY19}. When random searching a mixed dimension scheme, we applied the same feature blocking as we did for \methodns. 
%, i.e., we sort features according to their frequency and divide them into multiple blocks. Features in each block share a same dimension. 
%Here we employ the same feature blocking scheme as \method for a fair comparison. 
Following the intuition that high-frequency features desire larger numbers of dimensions, we generated 16 random descending sequences as the search space of the mixed dimension scheme for each model and report the best results.

%the dimensions for feature blocks. We randomly generate 16 dimension schemes for each model and report the best results.\\
%随机搜索是NAS问题一个很强的baseline（引用）。在我们的实验中，我们做随机搜索时也将feature排序及分块，为了公平性，我们为包括随机搜索在内的不同方法采用了同样的分块方案；此外，我们优化了搜索空间，会产生一个最大值小于d0的随机的递减序列，作为每个块的特征维度（因为考虑到高频特征的维数应该更高）。我们同样为每个模型随机搜索了16组方案，并汇报效果最好的方案。
{$\bullet$ \bf MDE} (Mixed Dimension Embeddings~\cite{ginart2019mixed}). This method performs feature blocking and applies
% where we employ the same blocking scheme as \method for a fair comparison. MDE propose 
a heuristic scheme where the number of dimensions per feature block is proportional to some fractional power of its frequency. 
%The scheme is determined with two hyperparameters, which 
We tested 16 groups of hyperparameters settings as suggested in the original paper and report the best results.
% 这个方法也将feature排序分块，然后采用一种启发式的方案，根据特征块的popularity的指数来确定其特征维数，其中存在一个超参系数；我们这里也采用相同的分块方式，并测试了16个超参选项
%assigned a dimension proportional to some fractional power of its popularity.

{$\bullet$ \bf NIS-ME} (Neural Input Search with Multi-size Embedding~\cite{DBLP:conf/kdd/JoglekarLCXWAKL20}). This method uses reinforcement learning to find optimal embedding dimensions for different features within a given memory budget. Since the implementation is not available, we follow the same experimental settings as detailed in~\cite{DBLP:conf/kdd/JoglekarLCXWAKL20}) and report the results of our method for comparison.

For {\bf \methodns}, we show its performance before and after the dimension pruning in Equation (\ref{equ:pruneindex}), and report the storage size of the pruned sparse matrix $\mathbf{E'}$ using COO format of sparse matrix~\cite{2020SciPy-NMeth}. We provide the results with different compression rates (CR), i.e., the division of unpruned embedding parameter size by the pruned size.

\noindent{\bf Implementation details.}
We implement our method using Pytorch~\cite{DBLP:conf/nips/PaszkeGMLBCKLGA19}. We apply Adam optimizer with the learning rate of 0.001 for model parameters $\Theta$ and that of 0.01 for soft selection layer parameters $\bm{\alpha}$. The mini-batch size is set to 4096 and the uniform base dimension $K$ is set to 64 for all the models. We apply the same blocking scheme for Random Search, MDE and \methodns.
The default numbers of feature blocks $L$ is set to 10 and 6 for Movielens and Criteo datasets, respectively. 
We employ various latent factor models: MF, MLP~\cite{DBLP:conf/www/HeLZNHC17} and NeuMF~\cite{DBLP:conf/www/HeLZNHC17} for rating prediction, and FM~\cite{FM}, Wide\&Deep~\cite{DBLP:conf/recsys/Cheng0HSCAACCIA16}, DeepFM~\cite{DBLP:conf/ijcai/GuoTYLH17} for CTR prediction, where the configuration of latent factor models are the same over different methods for a fair comparison. Besides, we exploit early-stopping for all the methods according to the change of validation loss during model training. All the experiments were performed using NVIDIA GeForce RTX 2080Ti GPUs.

\subsection{Comparison Results}

\begin{table*}[t] 
\captionsetup{labelfont=bf}
	\centering 
	\caption{Comparison between \method and baselines on the rating prediction task using Movielens-20M dataset. We also report the storage size of the derived feature embeddings and the training time per method. For \methodns, we show its results with and w/o different compression rates (CR), i.e., the division of unpruned embedding parameter size by the pruned size.
%	, where the pruning rate (PR) is evaluated after pruning.
%		Here we also report the storage size of model's embedding layer and the total training time cost of each method. For \methodns, we further show its performance using parameters pruning, with the compression rate of 2 and 2.5.
}\label{tab:compare_cf} 
\resizebox{.9\linewidth}{!}{
		\begin{tabular}{l|ccc|ccc|ccc}
		\toprule[1pt]
		\multirow{3}{*}{\centering \bf{Search Methods}}&\multicolumn{3}{c}{MF}
		&\multicolumn{3}{|c|}{MLP}
		&\multicolumn{3}{c}{NeuMF}\\
		&Params&\multirow{2}{*}{\centering Time Cost}&\multirow{2}{*}{MSE}&Params&\multirow{2}{*}{\centering Time Cost}&\multirow{2}{*}{MSE}&Params&\multirow{2}{*}{\centering Time Cost}&\multirow{2}{*}{MSE}\\
		&(M)&&&(M)&&&(M)&\\
		\midrule[0.5pt]
		Grid Search&33&16$h$&0.622&35&8$h$&0.640&61&4$h$&0.625\\
		\midrule[0.5pt]
			Random Search&33&16$h$&0.6153&22&4$h$&0.6361&30&2$h$&\underline{0.6238}\\
		\midrule[0.5pt]
		MDE&35&24$h$&\underline{0.6138}&35&5$h$&\underline{0.6312}&27&3$h$&0.6249\\
		\midrule[0.5pt]
		DNIS (unpruned)&37&1$h$&\bf{0.6096}&36&1$h$&\bf{0.6255}&72&1$h$&\bf{0.6146}\\
		DNIS ($CR=2$)&21&1$h$&\underline{0.6126}&20&1$h$&\underline{0.6303}&40&1$h$&\underline{0.6169}\\
		DNIS ($CR=2.5$)&17&1$h$&0.6167&17&1$h$&0.6361&32&1$h$&0.6213\\
		\bottomrule[1pt]
	\end{tabular}  
}
\vspace{-0.05in}
\end{table*}
 
\begin{table*}[t]
\captionsetup{labelfont=bf}
	\centering
		\caption{Comparison between \method and baselines on the CTR prediction task using Criteo dataset. 
%			Here we also report the storage size of model's embedding layer and the total training time cost of each method. For \methodns, we further show its performance using parameters pruning, with the compression rate of 20 and 30.
		}\label{tab:compare_ctr}
\resizebox{.9\linewidth}{!}{
		\begin{tabular}{l|p{0.9cm}<{\centering}cp{0.9cm}<{\centering}p{0.95cm}<{\centering}|p{0.9cm}<{\centering}cp{0.9cm}<{\centering}p{0.95cm}<{\centering}|p{0.9cm}<{\centering}cp{0.9cm}<{\centering}p{0.95cm}<{\centering}}
		\toprule[1pt]
		\multirow{3}{*}{\centering \bf{Search Methods}}&\multicolumn{4}{c}{FM}
		&\multicolumn{4}{|c|}{Wide\&Deep}
		&\multicolumn{4}{c}{DeepFM}\\
		&Params&\multirow{2}{0.6cm}{\centering Time Cost}&\multirow{2}{0.8cm}{AUC}&\multirow{2}{*}{Logloss}&Params&\multirow{2}{0.6cm}{\centering Time Cost}&\multirow{2}{*}{AUC}&\multirow{2}{*}{Logloss}&Params&\multirow{2}{0.6cm}{\centering Time Cost}&\multirow{2}{*}{AUC}&\multirow{2}{*}{Logloss}\\
		&(M)&&&&(M)&&&&(M)&\\
		\midrule[0.5pt]
		Grid Search&441&16$h$&0.7987&0.4525&254&16$h$&0.8079&0.4435&382&14$h$&0.8080&0.4435\\
		\midrule[0.5pt]
		Random Search&73&12$h$&\underline{0.7997}&\underline{0.4518}&105&16$h$&\underline{0.8084}&\underline{0.4434}&105&12$h$&\underline{0.8084}&\underline{0.4434}\\
		\midrule[0.5pt]
		MDE&397&16$h$&0.7986&0.4530&196&16$h$&0.8076&0.4439&396&16$h$&0.8077&0.4438\\
		\midrule[0.5pt]
		DNIS (unpruned)&441&3$h$&\bf{0.8004}&\bf{0.4510}&395&3$h$&\bf{0.8088}&\bf{0.4429}&416&3$h$&\bf{0.8090}&\bf{0.4427}\\
		DNIS ($CR=20$)&26&3$h$&\underline{0.8004}&\underline{0.4510}&29&3$h$&\underline{0.8087}&\underline{0.4430}&29&3$h$&\underline{0.8088}&\underline{0.4428}\\
		DNIS ($CR=30$)&17&3$h$&0.8004&0.4510&19&3$h$&0.8085&0.4432&20&3$h$&0.8086&0.4430\\
		\bottomrule[1pt]
	\end{tabular}  
} 
\vspace{-0.1in}
\end{table*}

\begin{table*}[t]
\captionsetup{labelfont=bf}
	\centering
		\caption{Comparison between \method and NIS-ME on Movielens-1M dataset. NIS-SE is a variant of NIS-ME method with a shared number of embedding dimension. Here we use the results of the original paper~\cite{DBLP:conf/kdd/JoglekarLCXWAKL20}.
		}\label{tab:compare_nis}
\resizebox{.9\linewidth}{!}{
		\begin{tabular}{c|cc|ccccccc}
		\toprule[1pt]
		Model&$K$&\#Params&Recall@1&Recall@5&@Recall@10&MRR@5&MRR@10&NDCG@5&NDCG@10\\
		\midrule[0.5pt]
				NIS-SE&16&390k&9.32&35.70&\underline{55.31}&18.22&20.83&22.43&28.63\\
				NIS-ME&16&390k&\underline{9.41}&\bf{35.90}&\bf{55.68}&\underline{18.31}&\underline{20.95}&\underline{22.60}&\bf{28.93}\\
				DNIS ($CR=2$)&16&390k&\bf{11.39}&\underline{35.79}&51.74&\bf{19.82}&\bf{21.93}&\bf{23.77}&\underline{28.90}\\ 
				\midrule[0.5pt]

						NIS-SE&16&195k&8.42&31.37&\underline{50.30}&15.04&17.57&19.59&25.12\\
						NIS-ME&16&195k&\underline{8.57}&\underline{33.29}&\bf{52.91}&\underline{16.78}&\underline{19.37}&\underline{20.83}&\underline{27.15}\\
				DNIS ($CR=4$)&16&195k&\bf{11.15}&\bf{33.34}&49.62&\bf{18.74}&\bf{20.88}&\bf{22.35}&\bf{27.58}\\
	\midrule[0.5pt]
				NIS-SE&32&780k&9.90&37.50&\underline{55.69}&19.18&21.60&23.70&29.58\\
				NIS-ME&32&780k&\underline{10.40}&\bf{38.66}&\bf{57.02}&\underline{19.59}&\underline{22.18}&\underline{24.28}&\underline{30.60}\\
				DNIS ($CR=2$)&32&780k&\bf{13.01}&\underline{38.26}&55.43&\bf{21.83}&\bf{24.12}&\bf{25.89}&\bf{31.45}\\
				\midrule[0.5pt]
						NIS-SE&32&390k&9.79&34.84&\underline{53.23}&17.85&20.26&22.00&27.91\\
				
						NIS-ME&32&390k&\underline{10.19}&\bf{37.44}&\bf{56.62}&\underline{19.56}&\underline{22.09}&\underline{23.98}&\bf{30.14}\\
				DNIS ($CR=4$)&32&390k&\bf{11.95}&\underline{35.98}&51.92&\bf{20.27}&\bf{22.39}&\bf{24.15}&\underline{29.30}\\		
		\bottomrule[1pt]
	\end{tabular}  
} 
\vspace{-0.1in}
\end{table*}

Table~\ref{tab:compare_cf} and Table~\ref{tab:compare_ctr} show the comparison results of different NIS methods on rating prediction and CTR prediction tasks, respectively. 
First, we can see that \method achieves the best prediction performance over all the model architectures for both tasks. 
It is worth noticing that the time cost of \method is reduced by $2\times$ to over $10\times$ compared with the baselines.  
%Specifically, \method outperforms the second-best model on Movielens with a decrease of 0.0054 on MSE on average, and on Criteo with an increase of 0.0006 on AUC on average. 
The results confirms that \method is able to learn discriminative feature embeddings with significantly higher efficiency than the existing search methods. 
Second, \method with dimension pruning achieves competitive or better performance than baselines, and can yield a significant reduction on embedding parameter size.
For example, \method with the CR of 2 outperforms all the baselines on Movielens, and yet reaches the minimal parameter size. The advantages of \method with the CR of 20 and 30 are more significant on Criteo. 
%and 30 outperforms all the baselines with the minimal parameter size on Movielens and Criteo datasets, respectively. 
Besides, we observe that \method can achieve a higher CR on Criteo than Movielens without sacrificing prediction performance. This is because the distribution of feature frequency on Criteo is severely skewed, leading to a significantly large number of redundant dimensions for low-frequency features. 
Third, among all the baselines, MDE performs the best on Movielens and Random Search performs the best on Criteo, while Grid Search gets the worst results on both tasks. This verifies the importance of applying mixed dimension embeddings to latent factor models.
% Note that all of the three baselines have searched over 16 groups of feature dimensions, and their time costs are slightly different due to the early-stopping of model training.
Fourth, we find that MF achieves better prediction performance on the rating prediction task than the other two model architectures. The reason may be the overfitting problem of MLP and NeuMF that results in poor generalization. Besides, DeepFM show the best results on the CTR prediction task, suggesting that the ensemble of DNN and FM is beneficial to improving CTR prediction accuracy.

Table~\ref{tab:compare_nis} shows the performance of \method and NIS-ME with respect to base embedding dimension $K$ and embedding parameter size for top-k item recommendation. From the results, we can see that \method achieves the best performance on most metrics (on average, the relative improvement over NIS-ME on Recall, MRR, and NDCG are $5.3\%$, $7.2\%$, and $2.7\%$, respectively). This indicates the effectiveness of \method by searching for the optimal mixed dimension scheme in a differentiable manner. Besides, NIS-ME shows consistent improvements over NIS-SE, admitting the benefit of replacing single embedding dimension with mixed dimensions. 
 
\begin{figure}
	\centering
	\vspace{-0.05in}
	\subcaptionbox{MSE \emph{vs} $K$.\label{fig:hyper:a}}
	[.49\linewidth]{\includegraphics[width=.49\linewidth]{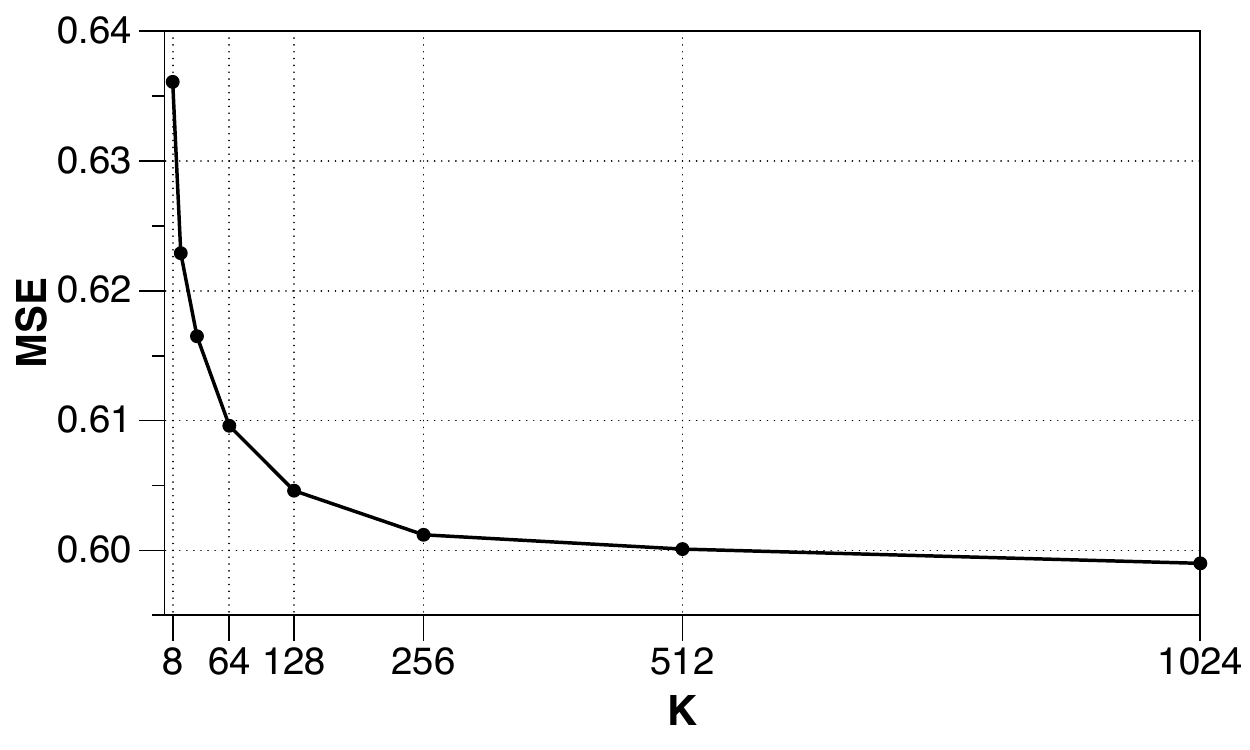}}
	\vspace{-0.05in}
	\subcaptionbox{MSE \emph{vs} $L$.\label{fig:hyper:b}}
	[.49\linewidth]{\includegraphics[width=.49\linewidth]{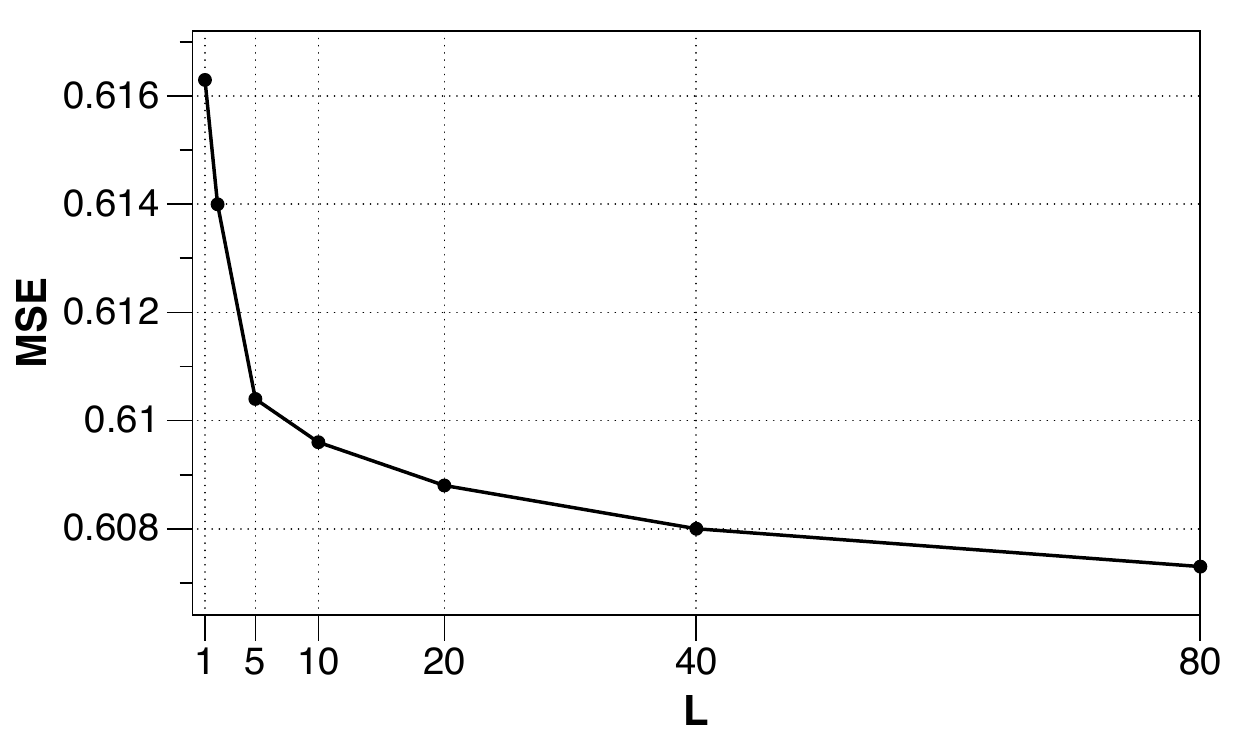}}
%		\vspace{-0.05in}
	\caption{Effects of hyperparameters on the performance of \methodns. We report the MSE results of MF on Movielens-20M dataset {w.r.t.} different base embedding dimensions $K$ and feature block numbers $L$.}\label{fig:hyper}
	\vspace{-0.1in}
\end{figure}

\begin{figure*}[t] 
	\centering 
	\subcaptionbox{${\bm{\alpha}}$ in different feature blocks.\label{fig:analysis:a}} 
	[.32\linewidth]{\includegraphics[width=.32\linewidth]{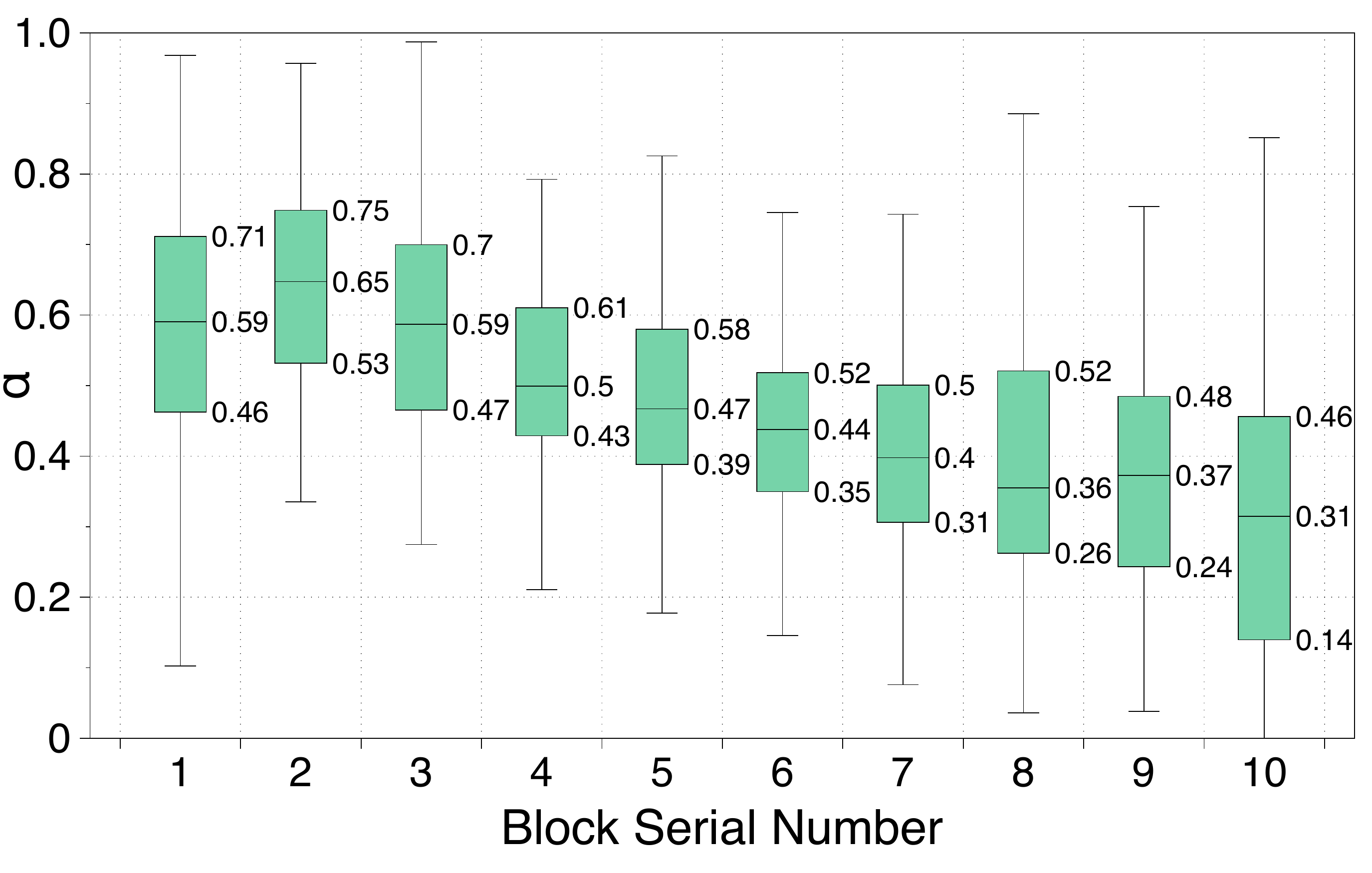}}
	\subcaptionbox{Embedding dimension \emph{vs} $\eta_f$.\label{fig:analysis:b}}
	[.32\linewidth]{\includegraphics[width=.32\linewidth]{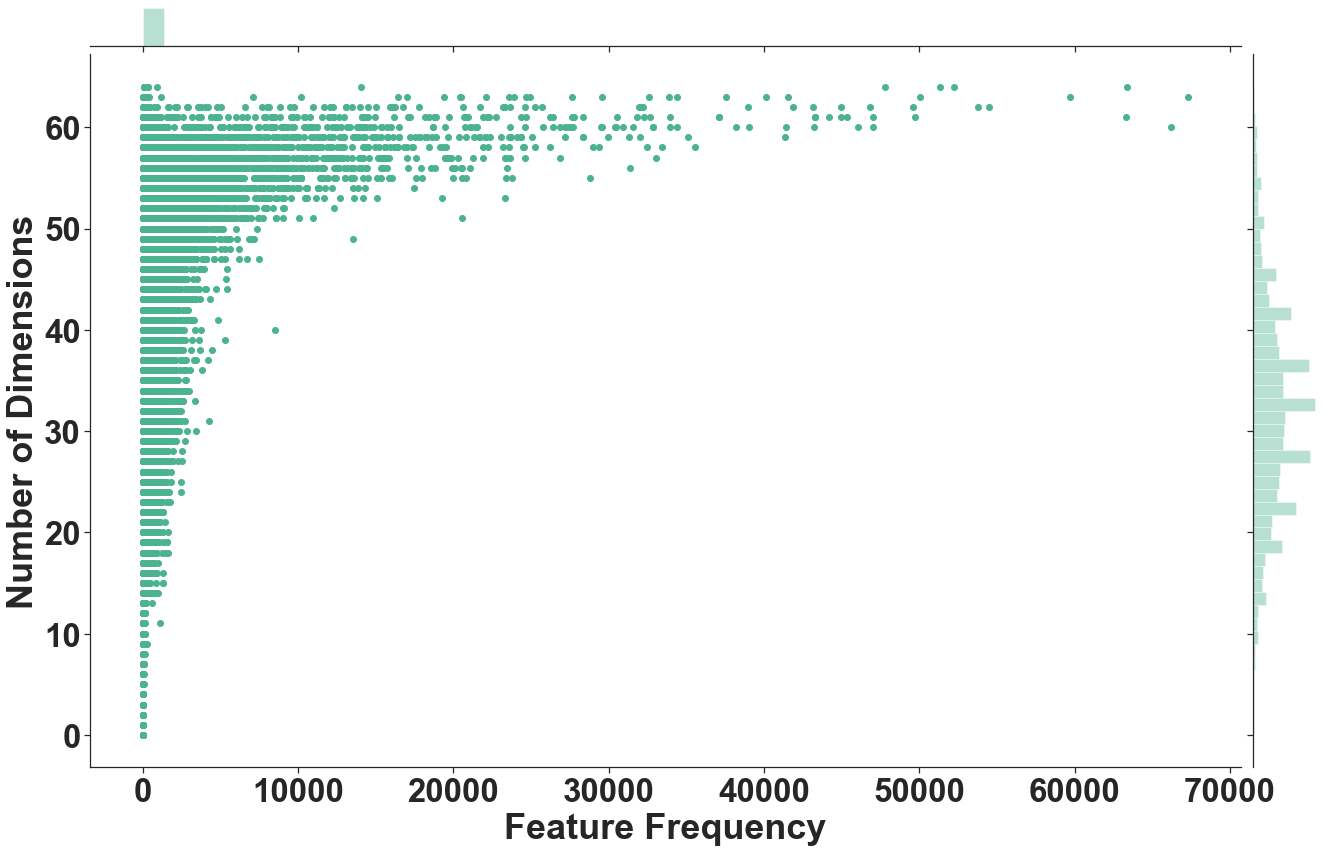}}
	 	\subcaptionbox{Performance \emph{vs} Pruning rates.\label{fig:analysis:c}}
	[.32\linewidth] {\includegraphics[width=0.32\linewidth]{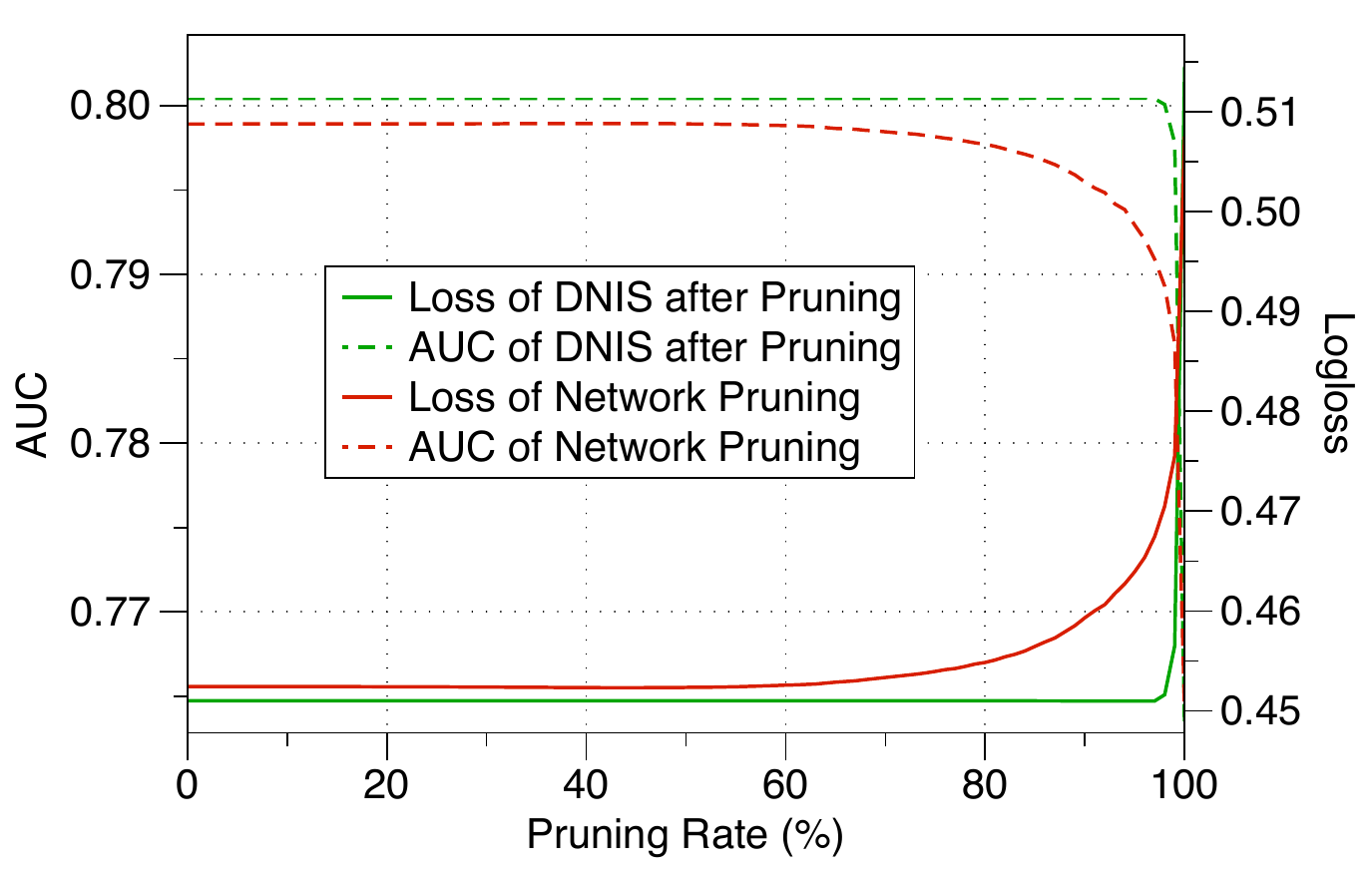}}
	\caption{(a) The distribution of trained parameters ${\bm{\alpha}}$ of the soft selection layer. Here we show the result of MF on Movielens dataset, where $L$ is set to 10. (b) The joint distribution plot of feature embedding dimensions and feature frequencies after dimension pruning. (c) Comparison of \method and network pruning performance over different pruning rates.}\label{fig:analysis}
\vspace{-0.1in} 
\end{figure*}

\vspace{-0.1in} 
\subsection{Hyperparameter Investigation}

We investigate the effects of two important hyperparameters $K$ and $L$ in \methodns. 
%The results are shown in Figure~\ref{fig:hyper}. 
Figure~\ref{fig:hyper:a} shows the performance change of MF {w.r.t.} different settings of $K$. We can see that increasing $K$ is beneficial to reducing MSE. This is because a larger $K$ allows a larger search space that could improve the representations of high-frequency features by giving more embedding dimensions. Besides, we observe a marginal decrease in performance gain. Specifically, the MSE is reduced by 0.005 when $K$ increases from 64 to 128, whereas the MSE reduction is merely 0.001 when $K$ changes from 512 to 1024. This implies that $K$ may have exceeded the largest number of dimensions required by all the features, leading to minor improvements.
Figure~\ref{fig:hyper:b} shows the effects of the number of feature blocks $L$. We find that increasing $L$ improves the prediction performance of \methodns, and the performance improvement decreases as $L$ becomes larger. 
% with a marginal decrease of performance gain for larger values of $k$. 
This is because dividing features into more blocks facilitates a finer-grained control on the embedding dimensions of different features, leading to more flexible mixed dimension schemes. 
%which then enhances feature representations.
Since both $K$ and $L$ affect the computation complexity of \methodns, %and interaction functions,
we suggest to choose reasonably large values for $K$ and $L$ to balance the computational efficiency and predictive performance based on the application requirements. 

\vspace{-0.1in}
\subsection{Analysis on DNIS Results}\label{section/compare_prune}

We first study the learned feature dimensions of \method through the learned soft selection layer ${\bm\alpha}$ and feature embedding dimensions after dimension pruning.
%To better understand the learned mixed dimension scheme, we visualize the parameters in the soft selection layer ${\bm\alpha}$ after training. 
Figure~\ref{fig:analysis:a} depicts the distributions of the trained parameters in $\bm\alpha$ for the 10 feature blocks on Movielens. Recall that the blocks are sorted in the descending order of feature frequency. 
%We first visualize the distribution of ${\bm{\alpha}}$ in Figure~\ref{fig:analysis:a}, which includes the parameters in the soft selection layer. Here we show the results of \method with MF model on Movielens, where we sort and split features into 10 blocks. Note the blocks are sorted in deceasing order of feature frequencies. 
%From the results, we can see that 
We can see that the learned parameters in ${\bm \alpha}$ for the feature blocks with lower frequencies converge to smaller values, indicating that lower-frequency features tend to be represented by smaller numbers of embedding dimensions. 
%Figure~\ref{fig:analysis:b} provides the number of embedding dimensions per feature after dimension pruning. 
Figure~\ref{fig:analysis:b} provides the correlation between embedding dimensions and feature frequency after dimension pruning. 
The results show that features with high frequencies end up with high embedding dimensions, whereas the dimensions are more likely to be pruned for low-frequency features. 
Nevertheless, there is no linear correlationship between the derived embedding dimension and the feature frequency. Note that the embedding dimensions for low-frequency features scatter over a long range of numbers. This is consistent with the inferior performance of MDE which directly determines the embedding dimensions of features according to their frequency.

We further compare \method with network pruning method~\cite{han2015learning}. For illustration purpose, we provide the results of the FM model on Criteo dataset. 
%shenyy: which network pruning method?
Figure~\ref{fig:analysis:c} shows the performance of two methods on different pruning rates (i.e., the ratio of pruned embedding weights). 
%shenyy: pruning rate or compression rate? be consistent!!!
From the result, \method achieves better AUC and Logloss results than network pruning over all the pruning rates. This is because \method optimizes feature embeddings with the gradients from the validation set, which benefits the selection of predictive dimensions, instead of simply removing redundant weights in the embeddings.

\vspace{-0.1in}
\section{Conclusion}

In this paper, we propose Differentiable Neural Input Search (\methodns), a method that searches for mixed features embedding dimensions in a differentiable manner through gradient descent.
The key idea of \method is to introduce a soft selection layer that controls the significance of each embedding dimension, and optimize this layer according to model's validation performance. 
We propose a gradient normalization technique and a fine-grained pruning procedure in \method to produce a flexible mixed embedding dimension scheme for different features.
%The key idea is to develop a soft dimension selection layer that controls the significance of each embedding dimension, and can be optimized with model's validation performance through gradient descent.
%We show that \method can be seamlessly incorporated with various existing latent factor models for recommendation.
The proposed method is model-agnostic, and can be incorporated with various existing architectures of latent factor models. 
We conduct experiments on three public real-world recommendation datasets. The results show that \method achieves the best predictive performance compared with existing neural input search methods with fewer embedding parameters and less time cost.
\bibliographystyle{named}
\bibliography{aaai21}

\begin{thebibliography}{42}
\providecommand{\natexlab}[1]{#1}
\providecommand{\url}[1]{\texttt{#1}}
\providecommand{\urlprefix}{URL }
\expandafter\ifx\csname urlstyle\endcsname\relax
  \providecommand{\doi}[1]{doi:\discretionary{}{}{}#1}\else
  \providecommand{\doi}{doi:\discretionary{}{}{}\begingroup
  \urlstyle{rm}\Url}\fi

\bibitem[{Baker et~al.(2017)Baker, Gupta, Naik, and
  Raskar}]{DBLP:conf/iclr/BakerGNR17}
Baker, B.; Gupta, O.; Naik, N.; and Raskar, R. 2017.
\newblock Designing Neural Network Architectures using Reinforcement Learning.
\newblock In \emph{5th International Conference on Learning Representations,
  {ICLR} 2017, Toulon, France, April 24-26, 2017, Conference Track
  Proceedings}.

\bibitem[{Bergstra, Yamins, and Cox(2013)}]{DBLP:conf/icml/BergstraYC13}
Bergstra, J.; Yamins, D.; and Cox, D.~D. 2013.
\newblock Making a Science of Model Search: Hyperparameter Optimization in
  Hundreds of Dimensions for Vision Architectures.
\newblock In \emph{Proceedings of the 30th International Conference on Machine
  Learning, {ICML} 2013, Atlanta, GA, USA, 16-21 June 2013}, 115--123.

\bibitem[{Cai, Zhu, and Han(2019)}]{DBLP:conf/iclr/CaiZH19}
Cai, H.; Zhu, L.; and Han, S. 2019.
\newblock ProxylessNAS: Direct Neural Architecture Search on Target Task and
  Hardware.
\newblock In \emph{7th International Conference on Learning Representations,
  {ICLR} 2019, New Orleans, LA, USA, May 6-9, 2019}.

\bibitem[{Chen et~al.(2018)Chen, Collins, Zhu, Papandreou, Zoph, Schroff, Adam,
  and Shlens}]{DBLP:conf/nips/ChenCZPZSAS18}
Chen, L.; Collins, M.~D.; Zhu, Y.; Papandreou, G.; Zoph, B.; Schroff, F.; Adam,
  H.; and Shlens, J. 2018.
\newblock Searching for Efficient Multi-Scale Architectures for Dense Image
  Prediction.
\newblock In \emph{Advances in Neural Information Processing Systems 31: Annual
  Conference on Neural Information Processing Systems 2018, NeurIPS 2018, 3-8
  December 2018, Montr{\'{e}}al, Canada}, 8713--8724.

\bibitem[{Chen et~al.(2019)Chen, Chen, He, Gao, Li, Lou, and
  Wang}]{DBLP:conf/kdd/ChenC0GLLW19}
Chen, Y.; Chen, B.; He, X.; Gao, C.; Li, Y.; Lou, J.; and Wang, Y. 2019.
\newblock {\(\lambda\)}Opt: Learn to Regularize Recommender Models in Finer
  Levels.
\newblock In \emph{Proceedings of the 25th {ACM} {SIGKDD} International
  Conference on Knowledge Discovery {\&} Data Mining, {KDD} 2019, Anchorage,
  AK, USA, August 4-8, 2019}, 978--986.

\bibitem[{Cheng et~al.(2016)Cheng, Koc, Harmsen, Shaked, Chandra, Aradhye,
  Anderson, Corrado, Chai, Ispir, Anil, Haque, Hong, Jain, Liu, and
  Shah}]{DBLP:conf/recsys/Cheng0HSCAACCIA16}
Cheng, H.; Koc, L.; Harmsen, J.; Shaked, T.; Chandra, T.; Aradhye, H.;
  Anderson, G.; Corrado, G.; Chai, W.; Ispir, M.; Anil, R.; Haque, Z.; Hong,
  L.; Jain, V.; Liu, X.; and Shah, H. 2016.
\newblock Wide {\&} Deep Learning for Recommender Systems.
\newblock In \emph{Proceedings of the 1st Workshop on Deep Learning for
  Recommender Systems, DLRS@RecSys 2016, Boston, MA, USA, September 15, 2016},
  7--10.

\bibitem[{Cheng et~al.(2018)Cheng, Shen, Zhu, and
  Huang}]{DBLP:conf/ijcai/ChengSZH18}
Cheng, W.; Shen, Y.; Zhu, Y.; and Huang, L. 2018.
\newblock {DELF:} {A} Dual-Embedding based Deep Latent Factor Model for
  Recommendation.
\newblock In \emph{IJCAI'18, July 13-19, 2018, Stockholm, Sweden}, 3329--3335.

\bibitem[{Cheng et~al.(2017)Cheng, Wang, Zhou, and Zhang}]{cheng2017survey}
Cheng, Y.; Wang, D.; Zhou, P.; and Zhang, T. 2017.
\newblock A survey of model compression and acceleration for deep neural
  networks.
\newblock \emph{arXiv preprint arXiv:1710.09282} .

\bibitem[{Colson, Marcotte, and Savard(2007)}]{DBLP:journals/anor/ColsonMS07}
Colson, B.; Marcotte, P.; and Savard, G. 2007.
\newblock An overview of bilevel optimization.
\newblock \emph{Annals {OR}} 153(1): 235--256.

\bibitem[{Covington, Adams, and Sargin(2016)}]{DBLP:conf/recsys/CovingtonAS16}
Covington, P.; Adams, J.; and Sargin, E. 2016.
\newblock Deep Neural Networks for YouTube Recommendations.
\newblock In \emph{Proceedings of the 10th {ACM} Conference on Recommender
  Systems, Boston, MA, USA, September 15-19, 2016}, 191--198.

\bibitem[{Domhan, Springenberg, and Hutter(2015)}]{DBLP:conf/ijcai/DomhanSH15}
Domhan, T.; Springenberg, J.~T.; and Hutter, F. 2015.
\newblock Speeding Up Automatic Hyperparameter Optimization of Deep Neural
  Networks by Extrapolation of Learning Curves.
\newblock In \emph{Proceedings of the Twenty-Fourth International Joint
  Conference on Artificial Intelligence, {IJCAI} 2015, Buenos Aires, Argentina,
  July 25-31, 2015}, 3460--3468.

\bibitem[{Elsken, Metzen, and Hutter(2019)}]{DBLP:conf/iclr/ElskenMH19}
Elsken, T.; Metzen, J.~H.; and Hutter, F. 2019.
\newblock Efficient Multi-Objective Neural Architecture Search via Lamarckian
  Evolution.
\newblock In \emph{7th International Conference on Learning Representations,
  {ICLR} 2019, New Orleans, LA, USA, May 6-9, 2019}.

\bibitem[{Franceschi et~al.(2018)Franceschi, Frasconi, Salzo, Grazzi, and
  Pontil}]{DBLP:conf/icml/FranceschiFSGP18}
Franceschi, L.; Frasconi, P.; Salzo, S.; Grazzi, R.; and Pontil, M. 2018.
\newblock Bilevel Programming for Hyperparameter Optimization and
  Meta-Learning.
\newblock In \emph{Proceedings of the 35th International Conference on Machine
  Learning, {ICML} 2018, Stockholmsm{\"{a}}ssan, Stockholm, Sweden, July 10-15,
  2018}, 1563--1572.

\bibitem[{Frankle and Carbin(2019)}]{DBLP:conf/iclr/FrankleC19}
Frankle, J.; and Carbin, M. 2019.
\newblock The Lottery Ticket Hypothesis: Finding Sparse, Trainable Neural
  Networks.
\newblock In \emph{7th International Conference on Learning Representations,
  {ICLR} 2019, New Orleans, LA, USA, May 6-9, 2019}.

\bibitem[{Ginart et~al.(2019)Ginart, Naumov, Mudigere, Yang, and
  Zou}]{ginart2019mixed}
Ginart, A.; Naumov, M.; Mudigere, D.; Yang, J.; and Zou, J. 2019.
\newblock Mixed Dimension Embeddings with Application to Memory-Efficient
  Recommendation Systems.
\newblock \emph{arXiv preprint arXiv:1909.11810} .

\bibitem[{Guo et~al.(2017)Guo, Tang, Ye, Li, and
  He}]{DBLP:conf/ijcai/GuoTYLH17}
Guo, H.; Tang, R.; Ye, Y.; Li, Z.; and He, X. 2017.
\newblock DeepFM: {A} Factorization-Machine based Neural Network for {CTR}
  Prediction.
\newblock In \emph{Proceedings of the Twenty-Sixth International Joint
  Conference on Artificial Intelligence, {IJCAI} 2017, Melbourne, Australia,
  August 19-25, 2017}, 1725--1731.

\bibitem[{Han et~al.(2015)Han, Pool, Tran, and Dally}]{han2015learning}
Han, S.; Pool, J.; Tran, J.; and Dally, W. 2015.
\newblock Learning both weights and connections for efficient neural network.
\newblock In \emph{Advances in neural information processing systems},
  1135--1143.

\bibitem[{Harper and Konstan(2016)}]{DBLP:journals/tiis/HarperK16}
Harper, F.~M.; and Konstan, J.~A. 2016.
\newblock The MovieLens Datasets: History and Context.
\newblock \emph{{ACM} Trans. Interact. Intell. Syst.} 5(4): 19:1--19:19.

\bibitem[{He et~al.(2017)He, Liao, Zhang, Nie, Hu, and
  Chua}]{DBLP:conf/www/HeLZNHC17}
He, X.; Liao, L.; Zhang, H.; Nie, L.; Hu, X.; and Chua, T. 2017.
\newblock Neural Collaborative Filtering.
\newblock In \emph{Proceedings of the 26th International Conference on World
  Wide Web, {WWW} 2017, Perth, Australia, April 3-7, 2017}, 173--182.

\bibitem[{Joglekar et~al.(2020)Joglekar, Li, Chen, Xu, Wang, Adams, Khaitan,
  Liu, and Le}]{DBLP:conf/kdd/JoglekarLCXWAKL20}
Joglekar, M.~R.; Li, C.; Chen, M.; Xu, T.; Wang, X.; Adams, J.~K.; Khaitan, P.;
  Liu, J.; and Le, Q.~V. 2020.
\newblock Neural Input Search for Large Scale Recommendation Models.
\newblock In \emph{{KDD} '20: The 26th {ACM} {SIGKDD} Conference on Knowledge
  Discovery and Data Mining, Virtual Event, CA, USA, August 23-27, 2020},
  2387--2397.

\bibitem[{Labs(2014)}]{criteo}
Labs, C. 2014.
\newblock Kaggle Display Advertising Challenge Dataset,
  http://labs.criteo.com/2014/02/kaggle-display-advertising-challenge-dataset/.

\bibitem[{Li et~al.(2017)Li, Kadav, Durdanovic, Samet, and
  Graf}]{DBLP:conf/iclr/0022KDSG17}
Li, H.; Kadav, A.; Durdanovic, I.; Samet, H.; and Graf, H.~P. 2017.
\newblock Pruning Filters for Efficient ConvNets.
\newblock In \emph{5th International Conference on Learning Representations,
  {ICLR} 2017, Toulon, France, April 24-26, 2017, Conference Track
  Proceedings}.

\bibitem[{Li and Talwalkar(2019)}]{DBLP:conf/uai/LiT19}
Li, L.; and Talwalkar, A. 2019.
\newblock Random Search and Reproducibility for Neural Architecture Search.
\newblock In \emph{Proceedings of the Thirty-Fifth Conference on Uncertainty in
  Artificial Intelligence, {UAI} 2019, Tel Aviv, Israel, July 22-25, 2019},
  129.

\bibitem[{Lian et~al.(2018)Lian, Zhou, Zhang, Chen, Xie, and Sun}]{xDeepFM}
Lian, J.; Zhou, X.; Zhang, F.; Chen, Z.; Xie, X.; and Sun, G. 2018.
\newblock xDeepFM: Combining Explicit and Implicit Feature Interactions for
  Recommender Systems.
\newblock In \emph{KDD'18, London, UK, August 19-23, 2018}, 1754--1763.

\bibitem[{Liu, Simonyan, and Yang(2019)}]{DBLP:conf/iclr/LiuSY19}
Liu, H.; Simonyan, K.; and Yang, Y. 2019.
\newblock {DARTS:} Differentiable Architecture Search.
\newblock In \emph{7th International Conference on Learning Representations,
  {ICLR} 2019, New Orleans, LA, USA, May 6-9, 2019}.

\bibitem[{Liu et~al.(2019)Liu, Sun, Zhou, Huang, and
  Darrell}]{DBLP:conf/iclr/LiuSZHD19}
Liu, Z.; Sun, M.; Zhou, T.; Huang, G.; and Darrell, T. 2019.
\newblock Rethinking the Value of Network Pruning.
\newblock In \emph{7th International Conference on Learning Representations,
  {ICLR} 2019, New Orleans, LA, USA, May 6-9, 2019}.

\bibitem[{Maclaurin, Duvenaud, and Adams(2015)}]{DBLP:conf/icml/MaclaurinDA15}
Maclaurin, D.; Duvenaud, D.; and Adams, R.~P. 2015.
\newblock Gradient-based Hyperparameter Optimization through Reversible
  Learning.
\newblock In \emph{Proceedings of the 32nd International Conference on Machine
  Learning, {ICML} 2015, Lille, France, 6-11 July 2015}, 2113--2122.

\bibitem[{Mendoza et~al.(2016)Mendoza, Klein, Feurer, Springenberg, and
  Hutter}]{DBLP:conf/icml/MendozaKFSH16}
Mendoza, H.; Klein, A.; Feurer, M.; Springenberg, J.~T.; and Hutter, F. 2016.
\newblock Towards Automatically-Tuned Neural Networks.
\newblock In \emph{Proceedings of the 2016 Workshop on Automatic Machine
  Learning, AutoML 2016, co-located with 33rd International Conference on
  Machine Learning {(ICML} 2016), New York City, NY, USA, June 24, 2016},
  58--65.

\bibitem[{Miller, Todd, and Hegde(1989)}]{DBLP:conf/icga/MillerTH89}
Miller, G.~F.; Todd, P.~M.; and Hegde, S.~U. 1989.
\newblock Designing Neural Networks using Genetic Algorithms.
\newblock In \emph{Proceedings of the 3rd International Conference on Genetic
  Algorithms, George Mason University, Fairfax, Virginia, USA, June 1989},
  379--384.

\bibitem[{Molchanov et~al.(2017)Molchanov, Tyree, Karras, Aila, and
  Kautz}]{DBLP:conf/iclr/MolchanovTKAK17}
Molchanov, P.; Tyree, S.; Karras, T.; Aila, T.; and Kautz, J. 2017.
\newblock Pruning Convolutional Neural Networks for Resource Efficient
  Inference.
\newblock In \emph{5th International Conference on Learning Representations,
  {ICLR} 2017, Toulon, France, April 24-26, 2017, Conference Track
  Proceedings}.

\bibitem[{Park et~al.(2018)Park, Naumov, Basu, Deng, Kalaiah, Khudia, Law,
  Malani, Malevich, Nadathur et~al.}]{park2018deep}
Park, J.; Naumov, M.; Basu, P.; Deng, S.; Kalaiah, A.; Khudia, D.; Law, J.;
  Malani, P.; Malevich, A.; Nadathur, S.; et~al. 2018.
\newblock Deep learning inference in facebook data centers: Characterization,
  performance optimizations and hardware implications.
\newblock \emph{arXiv preprint arXiv:1811.09886} .

\bibitem[{Paszke et~al.(2019)Paszke, Gross, Massa, Lerer, Bradbury, Chanan,
  Killeen, Lin, Gimelshein, Antiga, Desmaison, K{\"{o}}pf, Yang, DeVito,
  Raison, Tejani, Chilamkurthy, Steiner, Fang, Bai, and
  Chintala}]{DBLP:conf/nips/PaszkeGMLBCKLGA19}
Paszke, A.; Gross, S.; Massa, F.; Lerer, A.; Bradbury, J.; Chanan, G.; Killeen,
  T.; Lin, Z.; Gimelshein, N.; Antiga, L.; Desmaison, A.; K{\"{o}}pf, A.; Yang,
  E.; DeVito, Z.; Raison, M.; Tejani, A.; Chilamkurthy, S.; Steiner, B.; Fang,
  L.; Bai, J.; and Chintala, S. 2019.
\newblock PyTorch: An Imperative Style, High-Performance Deep Learning Library.
\newblock In \emph{Advances in Neural Information Processing Systems 32: Annual
  Conference on Neural Information Processing Systems 2019, NeurIPS 2019, 8-14
  December 2019, Vancouver, BC, Canada}, 8024--8035.

\bibitem[{Pedregosa(2016)}]{DBLP:conf/icml/Pedregosa16}
Pedregosa, F. 2016.
\newblock Hyperparameter optimization with approximate gradient.
\newblock In \emph{Proceedings of the 33nd International Conference on Machine
  Learning, {ICML} 2016, New York City, NY, USA, June 19-24, 2016}, 737--746.

\bibitem[{Real et~al.(2019)Real, Aggarwal, Huang, and
  Le}]{DBLP:conf/aaai/RealAHL19}
Real, E.; Aggarwal, A.; Huang, Y.; and Le, Q.~V. 2019.
\newblock Regularized Evolution for Image Classifier Architecture Search.
\newblock In \emph{The Thirty-Third {AAAI} Conference on Artificial
  Intelligence, {AAAI} 2019, The Thirty-First Innovative Applications of
  Artificial Intelligence Conference, {IAAI} 2019, The Ninth {AAAI} Symposium
  on Educational Advances in Artificial Intelligence, {EAAI} 2019, Honolulu,
  Hawaii, USA, January 27 - February 1, 2019}, 4780--4789.

\bibitem[{Real et~al.(2017)Real, Moore, Selle, Saxena, Suematsu, Tan, Le, and
  Kurakin}]{DBLP:conf/icml/RealMSSSTLK17}
Real, E.; Moore, S.; Selle, A.; Saxena, S.; Suematsu, Y.~L.; Tan, J.; Le,
  Q.~V.; and Kurakin, A. 2017.
\newblock Large-Scale Evolution of Image Classifiers.
\newblock In \emph{Proceedings of the 34th International Conference on Machine
  Learning, {ICML} 2017, Sydney, NSW, Australia, 6-11 August 2017}, 2902--2911.

\bibitem[{Rendle(2010)}]{FM}
Rendle, S. 2010.
\newblock Factorization Machines.
\newblock In \emph{{ICDM} 2010, The 10th {IEEE} International Conference on
  Data Mining, Sydney, Australia, 14-17 December 2010}, 995--1000.

\bibitem[{{Virtanen} et~al.(2020){Virtanen}, {Gommers}, {Oliphant},
  {Haberland}, {Reddy}, {Cournapeau}, {Burovski}, {Peterson}, {Weckesser},
  {Bright}, {van der Walt}, {Brett}, {Wilson}, {Jarrod Millman}, {Mayorov},
  {Nelson}, {Jones}, {Kern}, {Larson}, {Carey}, {Polat}, {Feng}, {Moore}, {Vand
  erPlas}, {Laxalde}, {Perktold}, {Cimrman}, {Henriksen}, {Quintero}, {Harris},
  {Archibald}, {Ribeiro}, {Pedregosa}, {van Mulbregt}, and
  {Contributors}}]{2020SciPy-NMeth}
{Virtanen}, P.; {Gommers}, R.; {Oliphant}, T.~E.; {Haberland}, M.; {Reddy}, T.;
  {Cournapeau}, D.; {Burovski}, E.; {Peterson}, P.; {Weckesser}, W.; {Bright},
  J.; {van der Walt}, S.~J.; {Brett}, M.; {Wilson}, J.; {Jarrod Millman}, K.;
  {Mayorov}, N.; {Nelson}, A. R.~J.; {Jones}, E.; {Kern}, R.; {Larson}, E.;
  {Carey}, C.; {Polat}, {\.I}.; {Feng}, Y.; {Moore}, E.~W.; {Vand erPlas}, J.;
  {Laxalde}, D.; {Perktold}, J.; {Cimrman}, R.; {Henriksen}, I.; {Quintero},
  E.~A.; {Harris}, C.~R.; {Archibald}, A.~M.; {Ribeiro}, A.~H.; {Pedregosa},
  F.; {van Mulbregt}, P.; and {Contributors}, S. .~. 2020.
\newblock {SciPy 1.0: Fundamental Algorithms for Scientific Computing in
  Python}.
\newblock \emph{Nature Methods} 17: 261--272.

\bibitem[{Xie et~al.(2019)Xie, Zheng, Liu, and Lin}]{DBLP:conf/iclr/XieZLL19}
Xie, S.; Zheng, H.; Liu, C.; and Lin, L. 2019.
\newblock {SNAS:} stochastic neural architecture search.
\newblock In \emph{7th International Conference on Learning Representations,
  {ICLR} 2019, New Orleans, LA, USA, May 6-9, 2019}.

\bibitem[{Zhao et~al.(2020)Zhao, Wang, Chen, Zheng, Liu, and
  Tang}]{zhao2020autoemb}
Zhao, X.; Wang, C.; Chen, M.; Zheng, X.; Liu, X.; and Tang, J. 2020.
\newblock AutoEmb: Automated Embedding Dimensionality Search in Streaming
  Recommendations.
\newblock \emph{arXiv preprint arXiv:2002.11252} .

\bibitem[{Zhong et~al.(2018)Zhong, Yan, Wu, Shao, and
  Liu}]{DBLP:conf/cvpr/ZhongYWSL18}
Zhong, Z.; Yan, J.; Wu, W.; Shao, J.; and Liu, C. 2018.
\newblock Practical Block-Wise Neural Network Architecture Generation.
\newblock In \emph{2018 {IEEE} Conference on Computer Vision and Pattern
  Recognition, {CVPR} 2018, Salt Lake City, UT, USA, June 18-22, 2018},
  2423--2432.

\bibitem[{Zoph and Le(2017)}]{DBLP:conf/iclr/ZophL17}
Zoph, B.; and Le, Q.~V. 2017.
\newblock Neural Architecture Search with Reinforcement Learning.
\newblock In \emph{5th International Conference on Learning Representations,
  {ICLR} 2017, Toulon, France, April 24-26, 2017, Conference Track
  Proceedings}.

\bibitem[{Zoph et~al.(2018)Zoph, Vasudevan, Shlens, and
  Le}]{DBLP:conf/cvpr/ZophVSL18}
Zoph, B.; Vasudevan, V.; Shlens, J.; and Le, Q.~V. 2018.
\newblock Learning Transferable Architectures for Scalable Image Recognition.
\newblock In \emph{2018 {IEEE} Conference on Computer Vision and Pattern
  Recognition, {CVPR} 2018, Salt Lake City, UT, USA, June 18-22, 2018},
  8697--8710.

\end{thebibliography}
\end{document}